  \documentclass[review,3p,times,fleqn,11pt]{elsarticle}
\usepackage{graphicx}
\graphicspath{{./}{../}}
\usepackage{amssymb}
\usepackage{natbib}
\bibpunct[, ]{(}{)}{,}{a}{}{,}%
\usepackage{lineno}
\usepackage{url}
\usepackage{color}
\usepackage{amsmath}
\usepackage{algorithm}  
\usepackage{algorithmic} 
\usepackage{appendix}
\usepackage{booktabs}
\usepackage{setspace}
\usepackage{subfigure}
\usepackage{tabularx}
\usepackage{lscape}
\usepackage{multirow}
\usepackage{tabularx}
\usepackage{makecell}
\usepackage{orcidlink}
\usepackage{ragged2e}
\usepackage{graphicx}       
\usepackage{float} 
\usepackage{booktabs}
\usepackage{threeparttable}
\usepackage{makecell}
\usepackage{graphicx}
\usepackage{adjustbox}
\newcolumntype{Y}{>{\raggedright\arraybackslash}X} 
\newcommand{\mycomment}[1]{}

\journal{EJOR}
\begin{document}
	
	\begin{frontmatter}
		
		
		\title{Large Language Model–Driven Full-Component Evolution of Adaptive Large Neighborhood Search}

		\author[label1]{Shaohua Yu*~\orcidlink{0000-0002-2110-4621}}
		\ead{shaohua.yu@njust.edu.cn}
            \author[label2]{Tianyu Chen}
            \ead{125101022415@njust.edu.cn}
            \author[label2]{Linyan Liu}
            \ead{liulinyan@njust.edu.cn}
		\author[label3,label4]{Jakob Puchinger}
		\ead{jpuchinger@em-normandie.fr}

		\address[label1]{School of Intelligent Science and Technology, Nanjing University of Science and Technology, Nanjing 210094, China}
        \address[label2]{School of Mechanical Engineering, Nanjing University of Science and Technology, Nanjing 210094, China}
        \address[label4]{Université Paris-Saclay, CentraleSupélec, Laboratoire Génie Industriel, 91190, Gif-sur-Yvette, France}
		\address[label3]{EM Normandie Business School, Métis Lab, 92110, Clichy, France}

\begin{abstract}

Adaptive Large Neighborhood Search (ALNS) is a prominent metaheuristic and a widely adopted approach for production and logistics optimization. However, it has long relied on hand-crafted components built on expert experience, which makes development slow and costly to adapt to new problems. This paper proposes a closed-loop, large-language-model-driven evolutionary framework that decouples ALNS and automatically rebuilds all of its components. We break ALNS into seven key modules: destroy, repair, operator selection, weight update, initial solution construction, acceptance rule, and destroy-rate control, and evolve each module through a dedicated task. 
By incorporating the Multi-dimensional Archive of Phenotypic Elites mechanism, the framework maintains a multi-dimensional elite archive to simultaneously drive the evolution of solution quality and strategic diversity. 
\textcolor{black}{In addition, we design multiple mechanisms, including parallel and sequential multi-module evolution as well as single-expert-driven and multi-expert-driven evolution, to systematically evaluate the impact of different evolutionary paradigms on algorithm generation performance.} Evaluations on Traveling Salesman Problem and Capacitated Vehicle Routing Problem benchmarks demonstrate that evolved algorithms consistently outperform optimized classic ALNS baselines under both fixed-iteration and fixed-time limits. The framework also shows a degree of generalizability and cross-problem transferability.  Code analysis also uncovers several counterintuitive yet meaningful design patterns that emerged naturally during evolution, offering practical and theoretical insights for future ALNS design. Finally, comparisons across multiple language models highlight clear differences in their ability to support evolutionary algorithm design, helping guide model selection for real-world engineering use.

\end{abstract}
    \begin{keyword}
Large Language models in OR; Adaptive Large Neighborhood Search; Automated Algorithm Evolution; Traveling Salesman Problem; Capacitated Vehicle Routing Problem
    \end{keyword}
\end{frontmatter}

\section{Introduction}

Combinatorial optimization problems are everywhere in key modern industries such as logistics and supply chains, and electronic design automation for chips. How efficiently we can solve these problems directly affects operating cost and response speed. 
The Traveling Salesman Problem (TSP) and the Capacitated Vehicle Routing Problem (CVRP) and their variants are classic NP-hard problems and have long been seen as benchmark “touchstones” for testing optimization algorithms \citep{pop2024comprehensive,voigt2025review}. 
When facing large-scale engineering instances, metaheuristic algorithms that can often produce high-quality feasible solutions within practical computational budgets, have become one of the mainstream approaches for tackling such problems \citep{blum2011hybrid}. Among them, Adaptive Large Neighborhood Search (ALNS) stands out because it can rebuild solution structures through competing operators and dynamically schedule strategies through adaptive mechanisms, achieving excellent performance in complex domains such as the TSP and the vehicle routing problem (VRP) family \citep{ropke2006adaptive,voigt2025review}.

However, building traditional ALNS algorithms has long been limited by a “manual design bottleneck.” Implementing a high-performance algorithm depends heavily on domain experts’ prior knowledge and a tedious trial-and-error process. Designers not only need to handcraft operators tailored to a specific problem structure, but also have to carefully tune the weight-update formula, parameters in the acceptance rule, and the logic that controls the level of destruction, among many other details. This experience-driven trial-and-error paradigm leads to long development cycles, adapts poorly to changing environments, and is constrained by the designer’s own perspective, often failing to discover algorithmic logic that is non-intuitive yet computationally more effective.

\textcolor{black}{Before the recent rise of large language models (LLMs), the automated design of heuristics had already been explored in evolutionary computation, especially through genetic programming (GP) \citep{o2009riccardo} and hyper-heuristics \citep{burke2013hyper}. GP evolves program structures automatically through operations such as mutation and recombination, and has been used to generate heuristics or adaptive decision rules for combinatorial optimization. Related lines of work, such as GP-based and grammatical-evolution-based hyper-heuristics, have shown that components including heuristic selection rules, acceptance criteria, and neighborhood combinations can be automatically evolved rather than manually handcrafted. However, these approaches typically rely on predefined program representations and syntax-level search, which limits the extent of high-level semantic redesign across the full ALNS pipeline.}

In recent years, LLMs such as GPT, DeepSeek, and Grok have made major breakthroughs in code generation and logical reasoning, opening a new path to break through this algorithm design bottleneck. Studies including DeepMind’s FunSearch \citep{romera2024mathematical} and AlphaEvolve \citep{novikov2025alphaevolve}, as well as NVIDIA’s Eureka \citep{ma2023eureka}, suggest that LLMs may surpass human baselines in areas like mathematical discovery and strategy design. This signals a broader shift in algorithm design—from “expert-crafted by humans” to “automatically evolved with the help of large models.”

Even so, using LLMs to design complex meta-heuristics is still at an early, exploratory stage. Existing work, including ReEvo \citep{ye2024reevo} and LLM-based multiobjective evolutionary optimization \citep{liu2025large}, shows that LLMs can generate heuristic operators, but most of these efforts focus on local optimization at the “operator level,” while treating the “decision layer” (e.g., weight updates) and the “control layer” (e.g., acceptance criteria) as fixed. This kind of “unbalanced evolution” can easily create bottlenecks: even advanced heuristic operators are constrained by traditional decision logic (such as static weights or fixed thresholds), limiting overall performance due to the weakest link in the algorithmic chain. In ALNS, performance depends not only on how good the operators are, but also on whether the decision and control layers can adaptively match the operators’ search behavior. Therefore, what is urgently needed is a full-pipeline automated design framework that can systematically redesign everything from low-level operators to high-level control logic.

To address these challenges, this paper develops an LLM-driven full-component evolutionary framework for ALNS and evaluates it beyond a single operator or a single problem setting. Instead of only generating local destroy or repair heuristics, we decompose ALNS into seven interacting modules: initial-solution construction, destroy operators, repair operators, operator selection, weight update, acceptance criterion, and destruction-degree control. Each module is redesigned through a dedicated code-evolution task within a closed generate–evaluate–feedback loop. To maintain both solution quality and strategic diversity, the framework incorporates the Multi-dimensional Archive of Phenotypic Elites (MAP-Elites) mechanism \citep{Cully2018QDO}, while isolated and cascade evaluators are used to assess candidate components under controlled computational budgets. We further instantiate the framework along two methodological axes, namely parallel versus serial component scheduling and single-expert versus multi-expert prompt orchestration. This design allows us to examine not only whether LLM-evolved ALNS outperforms hand-crafted ALNS, but also how performance depends on component interactions, evolution paradigms, language-model choice, and cross-problem transfer. The resulting algorithms are evaluated on standard TSPLIB and CVRPLIB benchmarks under fixed-iteration, fixed-time, and extended-budget regimes.

The main contributions of this work are as follows:

\begin{enumerate}
    \item A full-component LLM-driven evolution framework for ALNS with clear empirical gains over tuned classic ALNS. We move beyond operator-level heuristic generation and propose a closed-loop framework that automatically evolves seven core ALNS components across the solution-operation, adaptive-decision, and global-control layers. By combining component-specific evolution tasks, MAP-Elites quality and diversity preservation, and cascade evaluation, the framework provides a systematic way to search over complete ALNS designs rather than isolated local heuristics. We validate the evolved ALNS variants on TSP and CVRP using TSPLIB and CVRPLIB benchmarks under fixed 1,000-iteration, fixed 60-second, and extended 10,000-iteration settings. Across these regimes, the evolved algorithms consistently achieve lower optimality gaps than a carefully tuned Baseline-ALNS, showing that LLM-driven evolution improves not only final solution quality but also practical computational efficiency, thereby shifting the empirical frontier defined by solution quality and time to the best-known solution (BKS).

    \item A systematic comparison of evolution paradigms and their compositional effects. We investigate four evolution regimes formed by parallel/serial component scheduling and single-/multi-expert prompt orchestration. The results show that the best paradigm is strongly problem- and budget-dependent. For TSP, serial strategies perform best under iteration‑based budgets, while a parallel strategy excels under strict time limits. For CVRP, parallel strategies emerge as the most competitive configurations in several budget scenarios, challenging the previous assumption that serial evolution is universally superior for constrained routing problems. We further reveal an important compositional phenomenon: components that perform well in isolation do not necessarily assemble into the best full algorithm, highlighting cross-component compatibility as a central issue in automated metaheuristic design.

    \item Scientific insights into LLM-driven algorithm design, model dependence, and transferability. By analyzing the evolved code and comparing multiple LLMs, we show that different models generate algorithms with distinct performance profiles: some are more effective under fixed iteration budgets, while others produce lower-overhead heuristics under strict time limits. Cross‑problem transfer experiments further indicate that part of the evolved high‑level ALNS logic is reusable, but transferability is highly asymmetric: CVRP‑evolved policies transfer to TSP with limited degradation and notable efficiency gains, while TSP‑evolved policies do not transfer robustly to CVRP.  The evolved code further reveals a coherent set of design principles: scale-normalized acceptance, non-monotone perturbation control, diversity-preserving adaptation, structure-aware operators, and moderately randomized reconstruction, that provide interpretable guidance for future ALNS design.
\end{enumerate}

The remainder of this paper is organized as follows: Section 2 introduces the related work, Section 3 is the methodology, Section 4 is numerical experiments, and Section 5 concludes.

\section{Related Work}

\subsection{Adaptive Large Neighborhood Search}

ALNS was introduced as a systematic extension of the Large Neighborhood Search (LNS) paradigm \citep{ropke2006adaptive}. Its key idea is to embed a competitive pool of operators and an adaptive scheduling mechanism into the classic destroy–repair loop. By adjusting operator selection probabilities based on their past performance, ALNS can automatically learn effective operator combinations during the search. As a result, it has become a widely used general framework for VRP and many of its variants. From a systems-engineering viewpoint, ALNS is often described as three tightly coupled layers: the search layer, the adaptive layer, and the control layer.

The search layer is the execution part of the algorithm. It includes initial solution construction and a set of destroy and repair operators. Typical destroy operators include Shaw Removal \citep{shaw1998using} and Worst Removal. Shaw Removal removes related customers based on spatio-temporal relatedness, while Worst Removal targets the most expensive elements in the current solution by identifying and removing edges (or assignments) that contribute most to the objective. On the repair side, a well-known example is Regret-k Insertion \citep{vidal2014unified}. Instead of inserting greedily, it measures the opportunity cost of delaying an insertion, which helps reduce the short-sightedness of standard greedy strategies. \cite{voigt2025review} provides detailed comparisons of different destroy and repair operators and summarizes their strengths and weaknesses.

The adaptive layer acts as the decision core \citep{turkevs2021meta}. It updates operator weights using historical feedback (such as how much the objective improves when an operator is applied) and then uses these weights to guide future operator selection. A common implementation is roulette-wheel selection, which converts weights into probabilities. Beyond the classic design, \cite{yu2025hybrid,yu2026new} proposed a new adaptive mechanism and reported clear improvements in solution quality through numerical experiments.

The control layer mainly defines the acceptance rule and how the neighborhood size is adjusted. A classic choice is simulated annealing, which accepts worse solutions with a certain probability to balance exploitation and exploration, reducing the risk of getting trapped in local optima. This layer often also includes a destruction degree controller to adjust the perturbation strength (i.e., neighborhood size) over time.

Although ALNS has been widely adopted, an important limitation remains: in many existing designs, “adaptiveness” is largely restricted to updating operator selection probabilities. Higher-level meta-strategies—such as the analytical form of the weight update rule, the reward design, the cooling schedule (or other parameter decay functions) in the acceptance criterion, and the logic for controlling destruction strength, are still commonly set by hand and hard-coded. This kind of static control can struggle on complex landscapes that are non-convex, multi-modal, and highly instance-dependent, which in turn limits robustness and generalization across instances.

\subsection{LLM-driven algorithm self-evolution and meta-optimization}

In recent years, intelligent paradigms for automatically designing algorithms have been developing rapidly. The core idea is to combine the logical reasoning and code-generation abilities of LLMs with evolutionary mechanisms inspired by natural selection, along with gradient-based and meta-optimization strategies. This is shifting algorithm design away from experience-based heuristics and manual tuning, toward a workflow where machines can “propose, test, and refine” interpretable algorithms and update rules on their own \citep{yu2025exploring}.

First, there is an LLM-driven evolutionary paradigm for discovering better algorithms. Representative methods such as FunSearch \citep{romera2024mathematical} and AlphaEvolve \citep{novikov2025alphaevolve} typically start with a population of programs that are functionally correct but not highly efficient (or candidates produced by an LLM). They then iterate through a loop of “selection–mutation–evaluation”:

\begin{itemize}
    \item Selection ranks programs by metrics such as runtime and resource consumption, and also considers diversity to choose a set of parent programs.
    \item Mutation uses an LLM as an “intelligent mutator,” rewriting code and recombining operators while preserving meaning and maintaining syntactic correctness, so the search can explore the algorithm space more effectively.
    \item Evaluation relies on automated testing frameworks to verify correctness on standard cases or in high-fidelity simulation, and to measure performance. Programs that pass verification and deliver meaningful improvements are added to a program library and reused as material for later rounds of evolution.
\end{itemize}

This paradigm usually produces human-readable programs or algorithm structures, which makes integration into real systems and safety auditing easier. It is therefore well suited to verified algorithm innovation in areas such as mathematical computation, system scheduling, and compiler optimization \citep{fawzi2022discovering}.

Second, there is meta-learning-based automatic discovery of reinforcement learning update rules. Instead of manually fixing temporal-difference targets or policy-gradient loss functions, this direction keeps the overall goal—maximizing long-term return—unchanged, but learns “how to learn” by building an update-rule meta-network \citep{oh2025discovering,jackson2024discovering}. The meta-network takes trajectory data collected from interacting with the environment as input, and outputs abstract predictive quantities and their target values. These outputs are then used to define how the agent updates its policy and internal representations. Multiple agents apply the same learned rule across different tasks, and their long-term returns serve as the outer objective for judging how good the rule is. By optimizing this outer objective via meta-gradients or evolutionary strategies across tasks and agents, the approach ultimately learns a unified update rule with stronger convergence and better generalization across a wide range of environments.

\textcolor{black}{Despite these important advances, existing self-evolving algorithm design approaches still differ substantially from the setting studied in this paper. Program-evolution frameworks such as FunSearch and AlphaEvolve typically optimize a monolithic program or a small number of routines under a unified evaluation interface, whereas meta-learning approaches in reinforcement learning mainly aim to discover learning rules for sequential decision-making agents.  In contrast, our focus is a modular metaheuristic for combinatorial optimization, namely ALNS, whose performance emerges from the interaction of heterogeneous components across the solution-operation, adaptive-decision, and global-control layers. Therefore, rather than evolving only a single operator or a single update rule, we propose a full-component evolutionary framework that decomposes ALNS into seven core modules and optimizes them under component-specific evaluation protocols. }

\section{Methodology}

\subsection{Overall Evolution Framework}

Traditional algorithm design often depends heavily on personal experience and repeated trial-and-error, which makes development slow and costly. To address this, this study builds an automated evolution framework driven by an LLM. The goal is to automatically generate and iteratively improve the components of an ALNS algorithm.

The framework follows a structured “generate–evaluate–feedback” loop. It leverages the LLM’s strength in understanding code and reasoning about logic, and lets the LLM act like an “intelligent mutation operator” that rewrites and improves algorithm components.

\begin{figure}[htbp]
    \centering
    \includegraphics[width=0.6\linewidth]{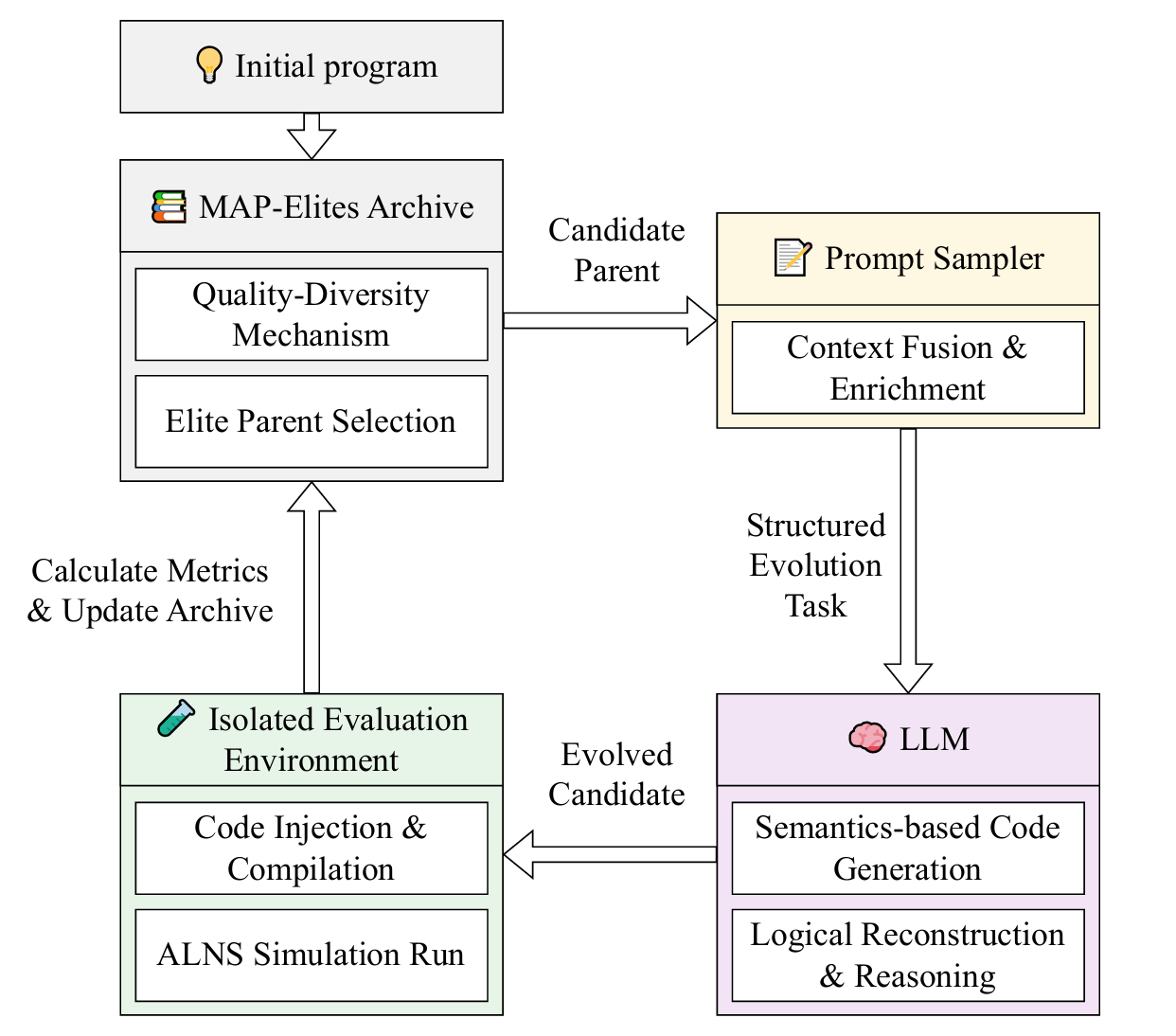}
    \caption{Closed-loop evolution process based on an LLM}
    \label{fig:1}
\end{figure}

The proposed framework, inspired by AlphaEvolve \citep{novikov2025alphaevolve}, operates as a closed loop with four key modules. The process starts with MAP-Elites, which samples parent candidates using a “quality and diversity” strategy and sends them to a Prompt Sampler. The sampler uses Context Fusion to dynamically assemble a structured evolution task, combining the parent code, historical execution feedback, and hard constraints. This prompt is then fed into the central LLM to produce new candidate code. Next, the generated code is placed into an isolated evaluation environment and tested with other components fixed as “sparring partners.” Finally, the test results are written back to the archive, completing one iteration (as shown in Figure \ref{fig:1}).

At the prompting level, this work does not use typical meta-prompt optimization. Instead, it adopts a “fixed instruction template + dynamic context injection” strategy: system instructions and hard constraints remain unchanged, so the LLM focuses its effort on improving the algorithm logic and code structure.

To evaluate the capability of the large language model to refine and innovate upon established optimization logic, the evolutionary process starts from a high-quality baseline comprising seven classic components of the adaptive large neighborhood search. The initial solution is generated through a nearest-neighbor greedy heuristic to provide a structured starting point. Within the search layer, the operator pool is initialized with well-known heuristics including random, similarity-based Shaw, and cost-based worst removal for the destroy phase, alongside greedy and regret-based strategies for the repair phase. The global control layer employs a fixed rule that removes fifty percent of the nodes to regulate destruction intensity and utilizes a standard simulated annealing criterion to govern solution acceptance. For the adaptive mechanism, the framework initializes with a classic roulette wheel operator selector and a weight updater that applies exponential smoothing to historical performance feedback. 

However, evolving a complete ALNS pipeline end-to-end creates an extremely large search space, which slows convergence. To address this issue, the study introduces a functionally decoupled framework that orthogonally decomposes ALNS into several cohesive core components that can be optimized independently. This decomposition substantially reduces the difficulty of evolution and improves search efficiency.  These components are grouped into three functional layers, which become the basis for independent evolution tasks:

\begin{itemize}
    \item Solution Operations: destroy operators, repair operators, and the initial solution generator. These directly act on the solution encoding and perform neighborhood transformations.
    \item Adaptive Mechanism: the operator selector and the weight updater. They allocate computation based on historical operator performance, reflecting online learning.
    \item Global Control Strategy: the acceptance rule and the destruction degree controller. They shape the convergence pace and exploration range, balancing exploitation and exploration.
\end{itemize}

With this decoupling, a complex algorithm-design problem becomes a set of clear and focused subproblems. This helps the LLM concentrate on improving one logical function at a time, while also making evolution more efficient under limited computing resources.

In the experimental section, the evolved components are integrated and validated through ablation studies to demonstrate the effectiveness of the proposed method.

\subsection{Component Specifications}

To facilitate full-component evolution, the ALNS framework is decomposed into seven distinct modules. Each module is treated as an independent evolutionary task with a functional interface designed to bridge high-level search logic with problem-specific operations. Table~\ref{tab:component_specs} summarizes the input context and the output requirements for each task. While the specific data structures (such as a single tour versus multiple routes) vary depending on the optimization problem, the functional role of each component within the ALNS cycle remains consistent.

\begin{table}[htbp]
    \centering
    \small
    \caption{Specifications for the Seven Evolved ALNS Components.}
    \label{tab:component_specs}
    \renewcommand{\arraystretch}{1.2}
    \begin{tabularx}{\textwidth}{l Y Y}
        \toprule
        \textbf{Component} & \textbf{Primary Input Context} & \textbf{Output Requirements} \\
        \midrule
        \multicolumn{3}{l}{\textit{\textbf{Layer 1: Solution Operators}}} \\
        Destroy Operator & Solution state, removal size $k$, instance data & Partial solution and a list of removed elements. \\
        Repair Operator & Partial solution, removed elements, instance data & A reconstructed feasible solution. \\
        Initial Solution Generator & Problem instance data (matrices, constraints) & A valid initial solution. \\
        \midrule
        \multicolumn{3}{l}{\textit{\textbf{Layer 2: Adaptive Mechanisms}}} \\
        Operator Selector & Operator weights and historical performance & Selected destroy and repair operator pair. \\
        Weight Updater & Performance feedback, iteration scores, lambda & Updated operator weight distribution. \\
        \midrule
        \multicolumn{3}{l}{\textit{\textbf{Layer 3: Global Controls}}} \\
        Acceptance Criterion & Objective costs, temperature, search progress & Boolean decision for solution replacement. \\
        Destruction Controller & Progress metrics and solution status & An integer $k$ for destruction intensity. \\
        \bottomrule
    \end{tabularx}
\end{table}

\subsection{Prompt Design}

In LLM-driven evolution, prompt design directly determines the quality and direction of the generated code. To support seven different evolution tasks efficiently and consistently, this study designs a general prompt template based on a "role–task–feedback–safety" structure.

To ensure the generated components are functionally correct and easy to integrate, all tasks follow a standardized instruction protocol:

\begin{itemize}
\item {Expert Role and Task Specificity}: Each prompt assigns a domain-expert persona (e.g., "expert in metaheuristic algorithms") and defines the module's role within the ALNS cycle.
\item {Structural Integrity via Diff Format}: The LLM is required to use a "Search/Replace" differential format.
\item Hard constraints: enforce strict rules, such as matching the required function signature, using only Python standard libraries and NumPy (no other third-party libraries), and wrapping code blocks with specific comment formats.
\end{itemize}

This standardized interface ensures that each generated component can be directly swapped into the ALNS main framework under consistent application programming interfaces.

Beyond the static template, each evolution prompt includes three top-performing programs based on their evaluation scores and two additional programs randomly selected from the remaining candidate pool. These reference programs are provided alongside their objective evaluation metrics and execution artifacts, including error logs and performance profiling data. This helps the LLM connect code logic with performance outcomes.

\subsection{Building the Evaluation System}
\label{S:BES}

Given limited computing resources and experiment time, this study sets up an efficient evaluation system based on controlled variables, aiming to measure the performance of ALNS components independently and accurately.

Because ALNS performance comes from nonlinear interactions among seven components, this work builds a dedicated Isolated Evaluator for each evolution task to remove interference between components. The evaluator uses a fixed-base design: when evolving one target component, all other components are held fixed. This provides a controlled environment and improves evaluation efficiency.

For quantitative scoring, a multi-objective quality model is constructed to evaluate the fitness of each candidate component across search effectiveness and computational efficiency. The aggregate metric, defined as the \textit{combined\_score}, is formulated as:
\begin{equation}
\text{Score}_{\text{combined}} = S_{quality} - P_{time} - P_{usage}
\end{equation}

The individual terms are calculated as follows:
\begin{itemize}
    \item Solution Quality ($S_{quality}$): Derived from the average gap relative to known optimal solutions across $N$ test instances. To normalize the score into a $[0, 1]$ range, the transformation is defined as:
    \begin{equation}
    S_{quality} = \frac{1}{1 + \overline{gap}}, \quad \overline{gap} = \frac{1}{N}\sum_{i=1}^{N}\frac{f_i - f^*_i}{f^*_i}
    \end{equation}
    where $f_i$ is the final cost obtained and $f^*_i$ is the best-known cost.
    
    \item Time Penalty ($P_{time}$): To prevent the evolution of computationally expensive logic, a time ratio $R_t = T_{avg} / T_{base}$ is monitored. The penalty is applied when $R_t$ exceeds a threshold $\tau$:
    \begin{equation}
    P_{time} = \max(0, (R_t - \tau) \cdot \lambda)
    \end{equation}
    The threshold $\tau$ is set to 1.2 and the penalty coefficient $\lambda$ is 0.01. This penalty is applied universally to all evolved components.
    
    \item Usage Penalty ($P_{usage}$): This penalty is specifically applied to the \textit{destroy} and \textit{repair} operator pools to ensure structural diversity within the ensemble. If the minimum selection frequency $\min(U)$ of any operator in the pool falls below a threshold $\rho$, a penalty is triggered:
    \begin{equation}
    P_{usage} = \max(0, (\rho - \min(U)) \cdot \eta)
    \end{equation}
    The parameters are configured as $\rho = 0.05$ and $\eta = 1.0$. For other components , $P_{usage}$ is set to zero.
\end{itemize}

To maintain population diversity, MAP-Elites is utilized to build an elite archive based on behavior feature space (behavior descriptors). This grid-based partitioning ensures the coexistence of candidates with distinct search patterns. For most components, a two-dimensional (2D) feature space is used: \textit{diversity} and \textit{stability}. The latter is calculated as $1 / (1 + \sigma(gaps))$. 

For components requiring more granular behavioral tracking, the feature space is expanded to three dimensions (3D) to preserve candidates across a broader spectrum of search patterns:
\begin{itemize}
    \item \textit{Operator Selector}: The dimensions include \textit{diversity}, \textit{stability}, and \textit{operator\_balance}. The \textit{operator\_balance} metric, calculated as the inverse of the standard deviation of operator usage, allows the archive to retain various selection logics, from highly specialized to broadly balanced, without directly penalizing the fitness score.
    \item \textit{Destruction Degree Controller}: The dimensions include \textit{diversity}, \textit{stability}, and \textit{avg\_k\_variance}. This configuration preserves various perturbation patterns, capturing both steady, incremental changes and highly volatile fluctuations in search-range control.
\end{itemize}

This customized design ensures each component is evolved more precisely on its key performance dimensions.

\subsection{Cascade Evaluation}
\label{S:SE}

To optimize computational resource allocation, an asynchronous multi-stage evaluation protocol is implemented. The protocol follows a hierarchical budget design with a 1:2 iteration ratio between the screening and assessment phases, ensuring a rapid filtering of low-quality candidates while providing sufficient search depth for promising ones. This protocol consists of two distinct phases:
\begin{itemize}
    \item Stage 1 (Screening Phase): Each candidate component is initially evaluated on small-scale instances with a budget of 500 iterations. A rank-based screening mechanism is applied: a candidate proceeds to the next stage only if its average gap ranks within the top 20\% of the total historical population of all previous evaluations in the current evolution run.
    \item Stage 2 (Full Assessment): Candidates passing Stage 1 undergo comprehensive evaluation across all benchmark instances with 1,000 iterations. This stage calculates the full suite of metrics, including the \textit{combined\_score}, stability, and behavioral descriptors required for the MAP-Elites archive.
\end{itemize}

\subsection{\textcolor{black}{Evolution Paradigms}}

\begin{figure}[htbp]
    \centering
    \subfigure[Component-scheduling modes]{\includegraphics[width=0.48\linewidth]{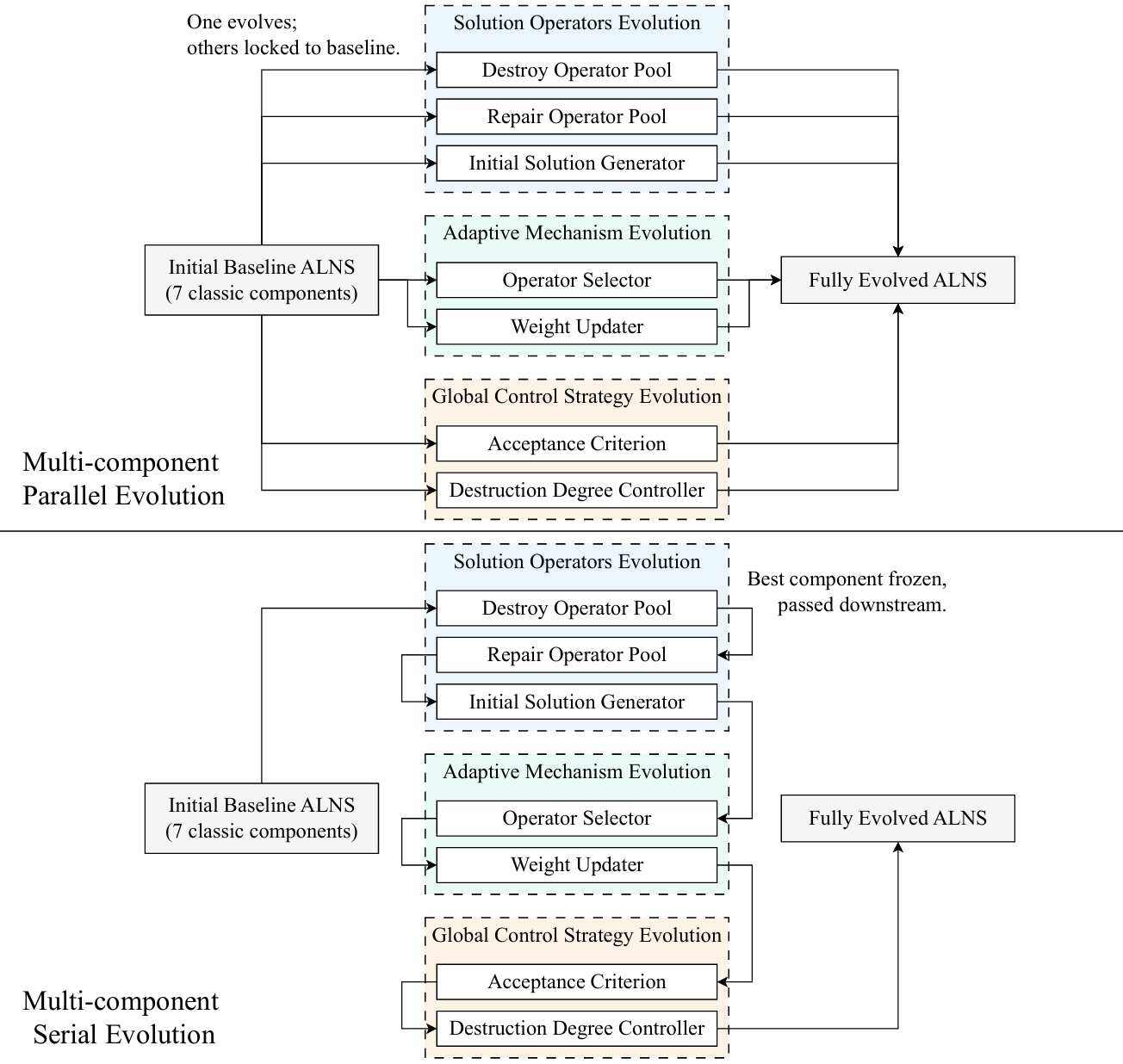}}
    \hfill
    \subfigure[Expert-orchestration modes]{\includegraphics[width=0.43\linewidth]{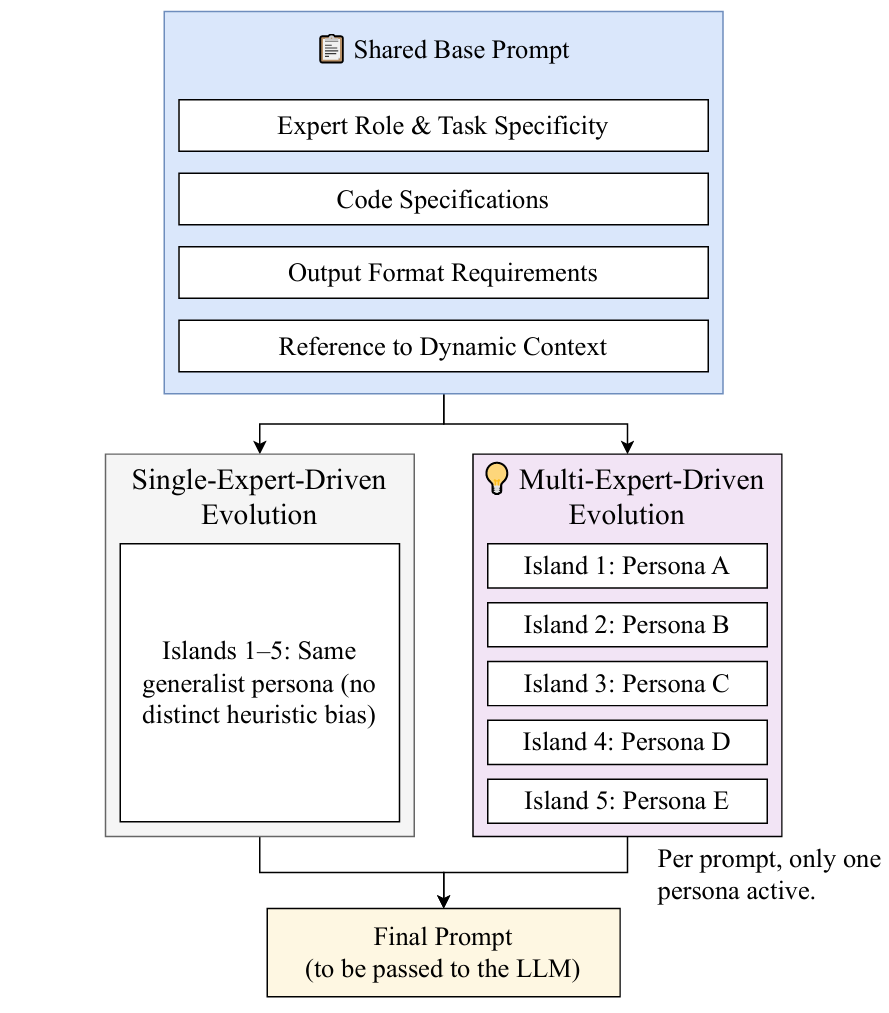}}
    \caption{The two methodological axes that define the four evolution paradigms studied in this paper.}
    \label{fig:evolution_modes}
\end{figure}

Beyond decomposing ALNS into seven modules, the proposed framework is instantiated along two orthogonal methodological axes: \emph{component scheduling} and \emph{expert orchestration}. The first axis determines whether the target components are evolved simultaneously or stage by stage; the second axis determines whether all evolution tasks share one common prompt persona or are guided by multiple differentiated expert personas. For compact reporting, we use Single-Expert and Multi-Expert to denote the two expert-orchestration modes. These two axes define four interpretable evolution regimes, summarized in Figure~\ref{fig:evolution_modes}. 

\subsubsection{Component-scheduling modes}
Let $\mathcal{C}=\{c_1,\ldots,c_K\}$ denote the set of ALNS components, where $K=7$ in this study. For each component $c_i$, let $\Omega_i$ be its candidate program space and let $F_i(\cdot)$ denote the isolated evaluation function defined in Section~\ref{S:BES}. Beyond the difference in optimization context, the two scheduling modes also differ substantially in wall-clock efficiency: serial evolution imposes hard precedence constraints because the evolution of one component can only begin after the previous component has finished and been frozen, whereas parallel evolution can launch all component-level searches simultaneously and is therefore much faster when sufficient computational resources are available.

\paragraph{Multi-component parallel evolution.}
In the parallel setting, each component is evolved as an independent task while all non-target components remain fixed to the standard baseline implementations specified in Table~\ref{tab:baseline-components-clean}. Formally, the best candidate for component $c_i$ is obtained as
\begin{equation}
x_i^{\star}=\arg\max_{x\in\Omega_i} F_i(x; B_{-i}),
\end{equation}
where $B_{-i}$ denotes the set of baseline implementations for all components except $c_i$. All $K$ component-evolution tasks are executed concurrently, and each task maintains its own MAP-Elites archive. After all tasks terminate, the best candidate from each archive is selected and assembled into a complete ALNS pipeline. Provided that adequate hardware resources are available, the wall-clock time of this setting is largely determined by the slowest single component-evolution task rather than by the sum of all task durations. The advantage of this setting is therefore high throughput and a stationary evaluation context; its limitation is that no direct selection pressure is imposed on cross-component compatibility during evolution.

\paragraph{Multi-component serial evolution.}
In the serial setting, components are optimized stage by stage according to a predefined order $\pi$. When evolving the component at stage $t$, the best evolved components from earlier stages are frozen and injected into the evaluator, while the remaining not-yet-evolved components stay at their baseline implementations. The stage-wise optimization is therefore defined as
\begin{equation}
x_{\pi(t)}^{\star}=\arg\max_{x\in\Omega_{\pi(t)}} F_{\pi(t)}(x; E_{<t}, B_{>t}),
\end{equation}
where $E_{<t}=\{x_{\pi(1)}^{\star},\ldots,x_{\pi(t-1)}^{\star}\}$ denotes the set of previously evolved components and $B_{>t}$ denotes the remaining baseline components. After each stage, the best candidate is frozen and passed forward to the next stage. In our implementation, the serial order follows the functional hierarchy of ALNS: initial solution, destroy operators, repair operators, operator selector, weight updater, acceptance criterion, and destruction-degree controller. Because each stage must wait for the previous one to finish, the total wall-clock cost accumulates across stages, making this mode substantially slower than parallel evolution in practice. Compared with the parallel setting, serial evolution partially exposes downstream components to an increasingly evolved algorithmic context and therefore captures part of the inter-component dependency.

\subsubsection{Expert-orchestration modes}
In this work, the term \emph{expert} refers to a role-conditioned prompt persona used to bias the reasoning style of the LLM, rather than to an additional human participant. The expert-orchestration axis controls how such personas are assigned during evolution, and therefore determines whether the search emphasizes a more focused and specialized evolutionary trajectory or a more diverse set of evolutionary trajectories.

\paragraph{Single-expert-driven evolution.}
In the single-expert setting, all component-evolution tasks and all evolutionary islands share the same generalist prompt persona. The instruction template, hard constraints, code-format requirements, and evaluation protocol are identical across tasks; only the dynamic task context differs from one component to another. This setting emphasizes a more focused and specialized evolutionary trajectory, encourages stylistic coherence across evolved modules, and reduces the risk that individually strong components become difficult to compose at the system level.

\paragraph{Multi-expert-driven evolution.}
In the multi-expert setting, the shared generalist persona is replaced by a predefined pool of expert personas $\mathcal{E}=\{e_1,\ldots,e_M\}$, such as operations-research, statistics, control, or physics-oriented roles. 
In the island-mapped implementation used here, each island keeps a fixed persona during one run, and candidate exchange occurs through elite migration rather than by altering persona assignments. This design broadens the diversity of possible evolutionary trajectories by introducing cognitive diversity at the prompt level.

\section{Numerical Experiments}

This section provides a systematic, multi-faceted evaluation of the proposed LLM-driven ALNS evolution framework. After describing the benchmark instances, comparison algorithms, parameter settings, computational budgets, and implementation details in Section~\ref{sec:setup}, the remainder of the chapter is organized around six research questions. We first ask whether LLM-driven evolution can outperform a carefully tuned, expert-designed ALNS baseline on TSP and CVRP under both fixed-iteration and fixed-time budgets (Section~\ref{sec:headline}), and then examine which of the four evolution paradigms, defined by the parallel/serial and single-/multi-expert axes, delivers the strongest performance and how that choice depends on the target problem (Section~\ref{sec:paradigm}). 
We next attribute the observed gains to specific modules through additive and subtractive ablation studies (Section~\ref{sec:attribution}), and examine whether the choice of the underlying LLM affects the quality of the evolved algorithms through a controlled cross-LLM comparison (Section~\ref{sec:llm}).
We then probe generalization by swapping high-level control components between TSP and CVRP to test the cross-problem transferability of the evolved logic (Section~\ref{sec:transfer}), \textcolor{black}{and benchmark the evolved ALNS against other recent LLM-driven and learning-based solvers (Section~\ref{sec:vs-evolved}).} Finally, Section~\ref{sec:discussion} discusses several effective design patterns that emerged during evolution. 

\subsection{Experimental Setup}
\label{sec:setup}

\subsubsection{Datasets}

We evaluate the proposed framework on instances drawn from the standard TSPLIB and CVRPLIB libraries. For both problems, the available instances are partitioned into an \emph{evolution set}, used during the LLM-driven evolution loop, and a \emph{test set}, used exclusively to assess generalization to unseen instances and to mitigate the risk that reported gains reflect memorization of specific instance patterns. 
The \emph{evolution set} is stratified by problem size into a small-scale (Small-Evo) and a medium-scale (Med-Evo) subset, and a screening subset of the smallest instances is used in Stage 1 of the cascade evaluator (Section \ref{S:SE}) to filter out low-quality candidates at low computational cost, while the full evolution set is used in Stage 2 for comprehensive scoring. The \emph{test set} is similarly stratified into a medium-small test subset (MS-Test) and a large-scale test subset (Lg-Test), with its largest instances serving to stress-test the scalability of the evolved heuristics beyond the regime seen during evolution. The complete list of instances in each category is provided in Appendix~\ref{A:1}.

\subsubsection{Algorithms under Comparison}
\label{S:Auc}

To quantitatively evaluate the automated algorithm design framework proposed in this work, we construct two core comparison algorithms: a baseline algorithm (Baseline-ALNS) and an evolved algorithm (Evolved-ALNS).

\textbf{Baseline-ALNS and Its Parameter Tuning Process}

Baseline-ALNS is a standard ALNS implementation designed by human experts, built from classic components widely used in the literature, see Table \ref{tab:baseline-components-clean}. It serves as a conventional and strong reference.

\begin{table}[thbp]
\centering
\caption{Component configuration of Baseline-ALNS.}
\label{tab:baseline-components-clean}
\renewcommand{\arraystretch}{1.2}
\begin{tabularx}{\linewidth}{p{0.25\linewidth} X}
\toprule
\textbf{Component} & \textbf{Strategy and description} \\
\midrule
\textbf{Destroy operators} 
    & $\bullet$ \textit{Random Destroy}: randomly remove $k$ nodes. \\
    & $\bullet$ \textit{Worst Destroy}: remove nodes yielding largest cost decrease. \\
    & $\bullet$ \textit{Shaw Destroy}: remove similar nodes based on relatedness. \\
\cmidrule(l){2-2} 

\textbf{Repair operators} 
    & $\bullet$ \textit{Greedy Repair}: reinsert at position with smallest cost increase. \\
    & $\bullet$ \textit{Regret-2 Repair}: insert based on the two largest regret values. \\
    & $\bullet$ \textit{Regret-3 Repair}: insert based on the three largest regret values. \\
\midrule

\textbf{Initial solution} & \textit{Greedy Initial Solution}: nearest-neighbor greedy construction. \\
\textbf{Operator selection} & \textit{Classic roulette wheel selection}. \\
\textbf{Weight update} & \textit{Classic weight updater}: discrete scores, exponential smoothing. \\
\textbf{Acceptance} & \textit{Classic acceptance criterion}: simulated annealing. \\
\textbf{Destruction controller} & Fixed destruction ratio. \\
\bottomrule
\end{tabularx}
\end{table}

To ensure a rigorous comparison, \textit{Baseline-ALNS} is tuned separately for TSP and CVRP via coordinate search on the evolution-set instances. Four hyperparameters are optimized sequentially---initial temperature $T_0$, cooling decay rate, weight smoothing reaction factor $\lambda$, and destruction ratio---with ten candidate values evaluated for each parameter while the others remain fixed at their current best settings. To reduce stochastic variance, each configuration is assessed through ten independent runs with fixed random seeds from 100 to 1000 under a budget of 1000 iterations. The resulting settings are $T_0 = 9000$, decay rate $= 0.9991$, $\lambda = 0.45$, and destruction ratio $= 0.5$ for TSP, and $T_0 = 7000$, decay rate $= 0.9994$, $\lambda = 0.85$, and destruction ratio $= 0.34$ for CVRP; these values are then used for all evolved variants and baseline comparisons on the corresponding problem.

\textbf{Evolved-ALNS and Evolutionary Framework Configuration} 

Evolved-ALNS is the main outcome of this work. It is generated entirely by the automated design framework LLM-ALNS-Evolve.
The framework decomposes ALNS into seven key components and evolves each component independently. After evolution, we select the best individual from each component's elite archive and integrate them into a complete algorithm (parallel evolution and serial evolution), yielding Evolved-ALNS.
In particular, the destroy and repair operators are chosen as the top three operators (by score) from the final elite archive of their respective evolution tasks.

During evolution, each candidate component is evaluated under the cascade evaluation. Specifically, each candidate first undergoes a 500-iteration screening evaluation on small-scale instances. Only candidates ranked in the top 20\% proceed to a 1,000-iteration full assessment on all evolution instances. This design balances evaluation reliability against the throughput required to assess a large number of candidates, and is supported by prior evidence that early-search behavior is strongly predictive of eventual convergence quality \citep{schede2025method}. 

Each component will evolve through 300 generations.  For each component, evolution was carried out using five independent islands in parallel, with an elite archive size of 150. The sampling temperature for the LLMs is maintained at 0.7 to balance stability and creative exploration. Elite migration was performed every 10 iterations with a migration rate of 10\%. Parent selection followed a biased strategy: with 70\% probability the process favored exploitation (mutation), and with 30\% probability it favored exploration (crossover/exploration).

Three large language models are employed as the evolutionary engines: \texttt{DeepSeek-V3.2}, \texttt{Grok-4.1-Fast}, and \texttt{Gemini-3-Flash}. Unless stated otherwise, all reported results are based on the variant evolved by \texttt{Gemini-3-Flash} (denoted Evolved-Gemini); the other two models are used exclusively in the cross-LLM comparison experiment.

\subsubsection{Evaluation Protocol }

To quantify solution quality, we adopt the relative percentage error (GAP) as the primary performance metric:
\begin{equation}
\mathrm{GAP}(\%) = \frac{f - f^{*}}{f^{*}} \times 100\%,
\end{equation}
where $f$ is the objective value obtained by the algorithm and $f^{*}$ is the known optimal (or best-known) value reported in TSPLIB / CVRPLIB.

To characterize performance from complementary angles, we evaluate algorithms under three budget scenarios:
\begin{enumerate}
\item \textbf{Fixed iterations (1{,}000).} Aligned with both the per-candidate evaluation budget during evolution and the per-configuration budget during baseline tuning, this scenario ensures no algorithm is implicitly favored by a budget mismatch between tuning/evolution and final assessment.
\item \textbf{Fixed runtime (60 seconds).} Mimics engineering deployment constraints and evaluates convergence efficiency under a strict wall-clock budget.
\item \textbf{Extended iterations (10{,}000).} Stress-tests whether the advantages observed at 1{,}000 iterations persist under an order-of-magnitude larger budget, probing long-run convergence beyond the regime seen during evolution.
\end{enumerate}

For all three scenarios, each algorithm is executed for ten independent runs per instance. Reported GAP and Runtime values represent the mean across these ten repetitions for each instance, and are subsequently aggregated by dataset split.

The computational budgets above serve as \emph{maximum allowable limits} for the search process. An early stopping criterion is consistently applied across all experimental stages: if the algorithm reaches the BKS of a given instance before the budget is exhausted, the execution terminates immediately, and the corresponding optimality gap is recorded as 0.0\%. The reported \textit{Runtime} for such instances reflects the actual time elapsed until BKS attainment rather than the nominal budget.

\subsubsection{Computing environment}

All experiments were run on a high-performance computing platform equipped with a 2.54~GHz AMD EPYC 7Y43 (48-core) CPU and 256~GB of memory, running a 64-bit Windows operating system. The code was implemented in Python~3.13.

\subsection{Does LLM-Driven Evolution Beat Hand-Crafted ALNS?}
\label{sec:headline}

We first address the central question of the paper: can LLM-driven evolution produce ALNS configurations that outperform a carefully tuned, expert-designed baseline? Table~\ref{tab:tsp_cvrp_evolution_modes} summarizes the answer across both problems (TSP and CVRP), three computational budgets (1{,}000 iterations, 10{,}000 iterations, and 60 seconds), and four dataset splits, yielding 24 problem--budget--split combinations for each evolution paradigm.

The overall pattern is clear and highly consistent. Across all 24 combinations, every evolved variant achieves a lower average optimality gap than the tuned Baseline-ALNS. In other words, the advantage of LLM-driven full-component evolution is not confined to a single problem, budget regime, or data split; it appears throughout the entire evaluation matrix.

The table also shows that the gains are not limited to solution quality, though the underlying mechanism varies by paradigm and problem. At fixed iteration budgets, the parallel variants complete iterations substantially faster than the baseline in many settings (e.g., on TSP Lg-Test, Parallel + SE reduces runtime from 4,721s to 217s), while the serial variants incur comparable per-iteration cost on several TSP splits. Under the fixed 60-second budget in many settings, TSP parallel variants exploit this speedup to run more iterations, whereas CVRP evolved variants typically run fewer but more effective iterations. Taken together, these results indicate that the evolved ALNS variants improve both search effectiveness and practical computational efficiency.

\providecommand{\subval}[1]{{\color{gray!70!black}\itshape #1}}
\providecommand{\cell}[2]{\makecell[r]{#1\\\subval{#2}}}
\providecommand{\cellbf}[2]{\makecell[r]{\textbf{#1}\\\subval{#2}}}

\begin{table}[!htbp]
\centering
\caption{Average optimality gap (\%) of evolution paradigms on TSP and CVRP benchmarks under three computational budgets.}
\label{tab:tsp_cvrp_evolution_modes}
\tiny
\setlength{\tabcolsep}{3pt}
\renewcommand{\arraystretch}{1.15}

\begin{tabular}{@{}ll
                rrrr
                rrrr
                rrrr@{}}
\toprule
 & & \multicolumn{4}{c}{\textbf{1{,}000 Iterations}}
   & \multicolumn{4}{c}{\textbf{10{,}000 Iterations}}
   & \multicolumn{4}{c}{\textbf{60 Seconds}} \\
\cmidrule(lr){3-6}\cmidrule(lr){7-10}\cmidrule(lr){11-14}
\textbf{Problem} & \textbf{Evolution Mode}
 & {Small-Evo} & {Med-Evo} & {MS-Test} & {Lg-Test}
 & {Small-Evo} & {Med-Evo} & {MS-Test} & {Lg-Test}
 & {Small-Evo} & {Med-Evo} & {MS-Test} & {Lg-Test} \\
\midrule

\multirow{5}{*}{\textbf{TSP}}
 & \textit{Baseline-ALNS}
   & \cell{1.714}{38.86s}   & \cell{2.007}{285.79s}  & \cell{2.749}{705.16s}  & \cell{4.177}{4{,}721.02s}
   & \cell{0.300}{266.78s}  & \cell{1.071}{2{,}842.05s}  & \cell{1.326}{6{,}965.08s} & \cell{3.251}{45{,}765.72s}
   & \cell{1.662}{1{,}176.1}  & \cell{3.723}{206.2}   & \cell{6.135}{147.1}   & \cell{10.378}{14.4} \\
 & Parallel + SE (default)
   & \cell{0.239}{3.89s}   & \cell{0.485}{24.72s}   & \cell{1.230}{45.99s}   & \cell{1.844}{216.69s}
   & \cell{0.047}{37.98s}   & \cell{0.161}{377.91s}   & \cell{0.493}{681.03s}   & \cell{1.026}{3423.82s}
   & \cellbf{0.098}{2{,}992}   & \cellbf{0.449}{2{,}275}   & \cellbf{1.138}{998}   & \cellbf{2.680}{299}
 \\
 & Parallel + ME
   & \cell{0.319}{13.71s}   & \cell{0.558}{73.39s}   & \cell{1.329}{140.98s}   & \cell{1.987}{640.92s}
   & \cell{0.086}{91.22s}   & \cell{0.204}{631.22s}   & \cell{0.559}{1263.85s}   & \cell{1.214}{5545.81s}
   & \cell{0.215}{3{,}744}   & \cell{0.608}{870}   & \cell{1.582}{679}   & \cell{3.729}{53}
 \\
 & Serial + SE
   & \cellbf{0.117}{33.6s}   & \cellbf{0.291}{340.49s}   & \cellbf{0.757}{775.35s}   & \cellbf{1.624}{4364.59s}
   & \cell{0.038}{255.34s}   & \cell{0.144}{4026.98s}   & \cell{0.492}{9009.49s}   & \cell{1.108}{59599.22s}
   & \cell{0.218}{498}   & \cell{0.760}{188}   & \cell{1.652}{131}   & \cell{3.900}{24}
 \\
 & Serial + ME
   & \cell{0.203}{27.7s}   & \cell{0.459}{306.34s}   & \cell{1.107}{789.11s}   & \cell{2.209}{4465.37s}
   & \cellbf{0.035}{177.49s}   & \cellbf{0.116}{2868.91s}   & \cellbf{0.323}{6758.24s}   & \cellbf{0.941}{44873.98s}
   & \cell{0.171}{941}   & \cell{0.791}{291}   & \cell{2.466}{211}   & \cell{4.461}{52}
 \\

\midrule

\multirow{5}{*}{\textbf{CVRP}}
 & \textit{Baseline-ALNS}
   & \cell{2.371}{3.17s}   & \cell{6.090}{22.84s}  & \cell{7.898}{31.49s}  & \cell{11.973}{412.61s}
   & \cell{0.851}{29.88s}  & \cell{2.653}{256.74s}   & \cell{2.894}{368.44s}   & \cell{4.753}{4{,}808.01s}
   & \cell{0.486}{9{,}452.1} & \cell{5.041}{3{,}254.7} & \cell{6.478}{4{,}517.2} & \cell{14.286}{152.3} \\
 & Parallel + SE (default)
   & \cell{0.206}{1.91s}   & \cell{1.049}{21.75s}   & \cellbf{1.152}{31.5s}   & \cellbf{2.739}{382.16s}
   & \cell{0.038}{11.33s}   & \cell{0.423}{251.11s}   & \cell{0.664}{360.33s}   & \cell{2.152}{5154.02s}
   & \cell{0.064}{1{,}763}   & \cell{0.859}{2{,}499}   & \cell{0.956}{3{,}348}   & \cellbf{3.732}{356}
 \\
 & Parallel + ME
   & \cell{0.544}{1.25s}   & \cellbf{1.027}{10.92s}   & \cell{1.187}{14.95s}   & \cell{3.044}{208.11s}
   & \cellbf{0.019}{6.36s}   & \cellbf{0.370}{161.39s}   & \cellbf{0.540}{241.98s}   & \cellbf{1.793}{3841.03s}
   & \cellbf{0.017}{1{,}924}   & \cellbf{0.641}{3{,}254}   & \cellbf{0.784}{3{,}788}   & \cell{4.304}{265}
 \\
 & Serial + SE
   & \cellbf{0.202}{1.1s}   & \cell{1.257}{14.69s}   & \cell{1.631}{19.06s}   & \cell{3.536}{272.24s}
   & \cell{0.038}{6.68s}   & \cell{0.514}{159.51s}   & \cell{0.772}{203.51s}   & \cell{2.366}{3160.25s}
   & \cell{0.039}{2{,}641}   & \cell{0.985}{4{,}178}   & \cell{1.411}{5{,}262}   & \cell{4.721}{253}
 \\
 & Serial + ME
   & \cell{0.302}{2.22s}   & \cell{1.218}{20.1s}   & \cell{1.367}{29.37s}   & \cell{3.082}{334.96s}
   & \cell{0.066}{13.31s}   & \cell{0.504}{203.96s}   & \cell{0.657}{304.37s}   & \cell{2.066}{3554.51s}
   & \cell{0.085}{2{,}420}   & \cell{0.934}{3{,}256}   & \cell{1.217}{3{,}282}   & \cell{4.025}{211}
 \\

\bottomrule
\end{tabular}

\vspace{0.2em}
 \begin{flushleft}\tiny
\textit{Notes.}
Each cell reports the average GAP (\%) on top and a secondary metric in italic gray below.
For the 1{,}000- and 10{,}000-iteration budgets the secondary metric is the average wall-clock runtime in seconds (s);
for the 60-second budget it is the average number of completed ALNS iterations.
Evolution modes: SE = Single-Expert prompt persona; ME = Multi-Expert prompt personas;
``default'' denotes the default configuration used throughout the main text.
Dataset splits: Small-Evo and Med-Evo (medium-scale) are used during the LLM-driven evolution loop;
MS-Test (medium-small) and Lg-Test (large-scale) are held out for generalization assessment.
Bold GAP values indicate the best (lowest) GAP per column among the four evolved variants;
the Baseline-ALNS row is excluded from the comparison.
All results are averaged over 10 independent runs per instance and aggregated by dataset split.
Runs that reach the best-known solution before exhausting the budget (The 1,000 iterations, 10,000 iterations, or 60 seconds) terminate early; reported runtimes reflect actual elapsed time in such cases.
\end{flushleft}
\end{table}

\begin{figure}[htbp]
    \centering
    \includegraphics[width=0.9\linewidth]{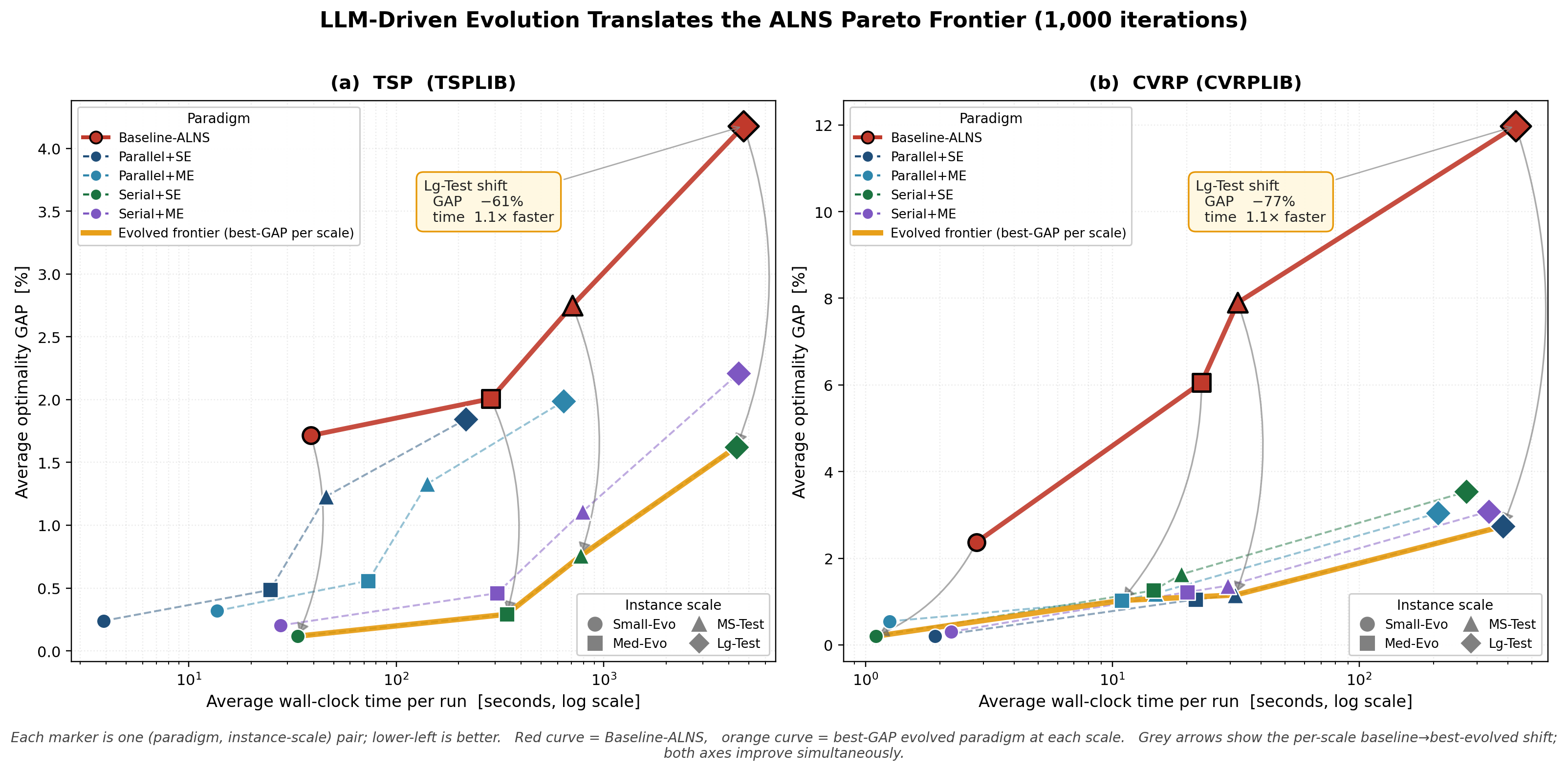}
    \caption{LLM-Driven Evolution Translates the ALNS Pareto Frontier (1,000 iterations)}
    \label{fig:ff1}
\end{figure}

To make this result more concrete, Figure~\ref{fig:ff1} provides a representative illustration of a phenomenon that is more important than a mere reduction in optimality gap. Specifically, it reveals a Pareto-like leap in which LLM-driven evolution improves solution quality and computational efficiency at the same time. In conventional algorithm engineering, more sophisticated and adaptive search logic is usually expected to produce better solutions only at the cost of higher per-iteration overhead. The evolved ALNS variants break this expectation. Rather than moving along the usual quality--speed trade-off curve, they shift the empirical frontier defined by solution quality and time to BKS toward the lower-left region of Figure~\ref{fig:ff1}, becoming both more accurate and faster simultaneously. The runtime statistics used in this comparison follow the evaluation protocol described in Section~\ref{sec:setup}.

\subsection{Which Evolution Paradigm Works Best?}
\label{sec:paradigm}

\begin{figure}[!htbp]
    \centering
    \includegraphics[width=0.9\linewidth]{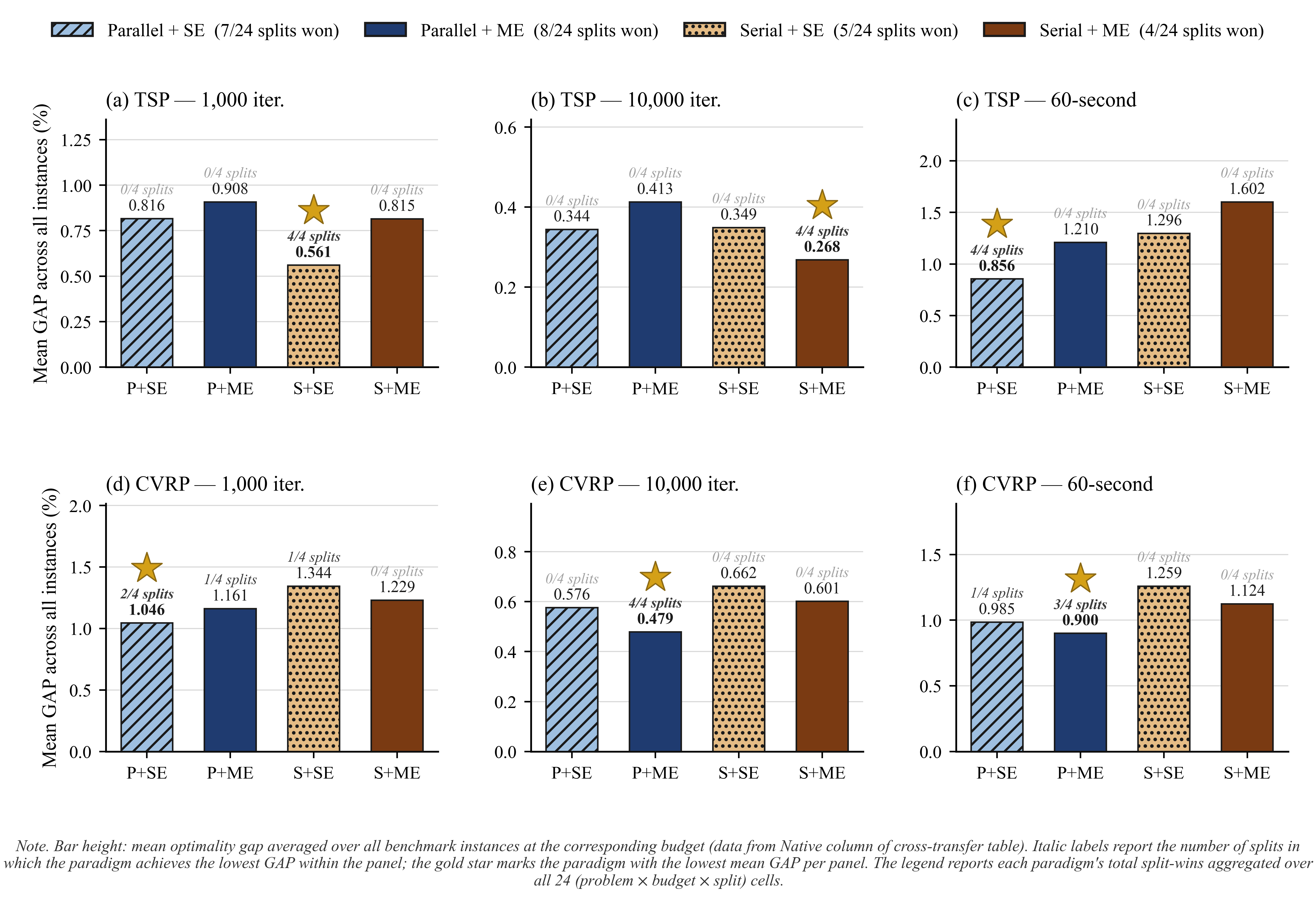}
    \caption{Macro-level comparison of evolution paradigms across problems and computational budgets.}
    \label{fig:fpc}
\end{figure}

This subsection compares the four evolution paradigms from two complementary angles. The purpose is not only to identify which setting performs best, but also to understand \emph{how} the differences appear: whether they are broad, high-level trends across problem--budget settings, or more fine-grained advantages that only become visible at the split level. We therefore first use Figure~\ref{fig:fpc} to give a macro-level view, and then return to Table~\ref{tab:tsp_cvrp_evolution_modes} for a more detailed micro-level comparison.

From the macro perspective in Figure~\ref{fig:fpc}, the main message is that no single evolution paradigm strictly dominates across all problems and budgets. On TSP, Serial + SE achieves the lowest mean GAP under 1,000 iterations, while Serial + ME performs best under 10,000 iterations; under the strict 60‑second budget, Parallel + SE remains the most efficient choice. 
On CVRP, the ranking is more volatile: Parallel + SE leads under 1,000 iterations, whereas Parallel + ME attains the best average performance under both 10,000 iterations and the 60-second budget.
The figure also makes the relative stability of each paradigm easier to read. The gold star marks the lowest mean GAP in each panel, while the italic win counts show how many of the four splits are won within that panel. This is useful because it separates two ideas that are easy to mix together: a paradigm may look strong on average, but that does not necessarily mean it wins consistently on every split.

At the finer‑grained level in Table~\ref{tab:tsp_cvrp_evolution_modes}, the differences are more nuanced. For TSP under 1{,}000 iterations, Serial + SE wins all four splits under TSP 1,000 iterations. Under 10{,}000 iterations, Serial + ME takes all four splits, with Serial + SE and Parallel + SE close behind; for example, Serial + ME reaches 0.941\% on Lg‑Test, while Parallel + SE achieves 1.026\%. Under the 60‑second budget, Parallel + SE again wins all four splits. 
On CVRP, the picture is less monolithic than previously observed. Under 1,000 iterations, Parallel + SE wins two of the four splits (MS-Test and Lg-Test), Parallel + ME wins Med-Evo, and Serial + SE wins Small-Evo. Under the 60-second budget, Parallel + ME wins three of the four splits (Small-Evo, Med-Evo, and MS-Test), while Parallel + SE wins Lg-Test. Under 10,000 iterations, Parallel + ME wins all four splits; Serial variants remain competitive but no longer dominate across every budget and split.
These observations reinforce the conclusion that the best evolution paradigm is both problem‑ and budget‑dependent, and that parallel strategies can now rival or surpass serial ones in several CVRP scenarios.

\subsection{Where Does the Improvement Come From? Component-Level Attribution}
\label{sec:attribution}

To investigate the exact sources of performance improvements within the evolutionary framework, we conduct comprehensive additive and subtractive ablation studies on all generated modules. These experiments span multiple evolution paradigms---Parallel versus Serial, and Single-Expert (SE) versus Multi-Expert (ME)---and are evaluated on the complete set of evolution instances under two computational budgets: a fixed iteration limit (1,000 iterations) and a fixed runtime limit (60 seconds). 

All ablation evaluations and baseline measurements were conducted in separate batches. Consequently, even for components that are logically identical to the baseline implementation, the measured performance under the fixed‑time budget (60 seconds) may display discrepancies due to uncontrolled environmental factors that differ across batches. 

\begin{table}[htbp]
\centering
\caption{Additive ablation results across evolution paradigms}
\label{tab:additive_ablation_modes}
\tiny
\setlength{\tabcolsep}{2.2pt}
\renewcommand{\arraystretch}{1.15}
\begin{threeparttable}
\resizebox{\textwidth}{!}{%
\begin{tabular}{llllllllll}
\toprule
Problem & Mode & Base GAP & Destroy & Repair & Initial & Selector & Weight & Acceptance & D. Ctrl. \\
\midrule
TSP & Parallel + SE 
& \makecell{1.848 \\ 2.607}
& \makecell{\textbf{0.487 (73.6)} \\ 0.997 (61.8)}
& \makecell{1.672 (9.5) \\ \textbf{0.792 (69.6)}}
& \makecell{1.112 (39.8) \\ 1.475 (43.4)}
& \makecell{1.384 (25.1) \\ 2.706 (-3.8)}
& \makecell{1.508 (18.4) \\ 2.653 (-1.8)}
& \makecell{1.280 (30.7) \\ 2.243 (14.0)}
& \makecell{0.806 (56.4) \\ 1.925 (26.2)}
\\

TSP & Parallel + ME 
& \makecell{1.848 \\ 2.607}
& \makecell{\textbf{0.464 (74.9)} \\ \textbf{0.640 (75.5)}}
& \makecell{1.632 (11.7) \\ 0.726 (72.2)}
& \makecell{0.898 (51.4) \\ 0.974 (62.6)}
& \makecell{1.366 (26.1) \\ 1.993 (23.6)}
& \makecell{1.450 (21.5) \\ 2.095 (19.6)}
& \makecell{1.340 (27.5) \\ 1.739 (33.3)}
& \makecell{0.631 (65.9) \\ 1.084 (58.4)}
\\

TSP & Serial + SE 
& \makecell{1.848 \\ 2.607}
& \makecell{\textbf{0.487 (73.6)} \\ \textbf{0.876 (66.4)}}
& \makecell{0.517 (72.0) \\ 0.909 (65.1)}
& \makecell{1.160 (37.2) \\ 1.363 (47.7)}
& \makecell{1.885 (-2.0) \\ 2.627 (-0.8)}
& \makecell{1.888 (-2.2) \\ 2.554 (2.0)}
& \makecell{1.663 (10.0) \\ 2.441 (6.4)}
& \makecell{1.147 (37.9) \\ 2.103 (19.3)}
\\

TSP & Serial + ME 
& \makecell{1.848 \\ 2.607}
& \makecell{\textbf{0.484 (73.8)} \\ \textbf{0.825 (68.4)}}
& \makecell{1.842 (0.3) \\ 2.220 (14.8)}
& \makecell{1.258 (31.9) \\ 1.422 (45.5)}
& \makecell{1.848 (0.0) \\ 2.411 (7.5)}
& \makecell{1.848 (0.0) \\ 2.414 (7.4)}
& \makecell{1.848 (0.0) \\ 2.407 (7.7)}
& \makecell{1.351 (26.9) \\ 1.756 (32.6)}
\\

\midrule

CVRP & Parallel + SE 
& \makecell{4.602 \\ 3.219}
& \makecell{4.558 (1.0) \\ 2.722 (15.4)}
& \makecell{\textbf{1.525 (66.9)} \\ \textbf{0.865 (73.1)}}
& \makecell{2.642 (42.6) \\ 1.773 (44.9)}
& \makecell{4.104 (10.8) \\ 2.589 (19.6)}
& \makecell{4.023 (12.6) \\ 2.435 (24.4)}
& \makecell{2.018 (56.2) \\ 1.890 (41.3)}
& \makecell{4.556 (1.0) \\ 2.788 (13.4)}
\\

CVRP & Parallel + ME 
& \makecell{4.602 \\ 3.219}
& \makecell{4.470 (2.9) \\ 2.633 (18.2)}
& \makecell{\textbf{1.439 (68.7)} \\ \textbf{0.839 (73.9)}}
& \makecell{2.930 (36.3) \\ 1.896 (41.1)}
& \makecell{3.983 (13.5) \\ 2.551 (20.8)}
& \makecell{3.970 (13.7) \\ 2.472 (23.2)}
& \makecell{2.029 (55.9) \\ 0.992 (69.2)}
& \makecell{4.456 (3.2) \\ 2.578 (19.9)}
\\

CVRP & Serial + SE 
& \makecell{4.602 \\ 3.219}
& \makecell{4.454 (3.2) \\ 2.482 (22.9)}
& \makecell{\textbf{1.366 (70.3)} \\ \textbf{0.778 (75.8)}}
& \makecell{3.482 (24.3) \\ 2.017 (37.3)}
& \makecell{4.311 (6.3) \\ 2.552 (20.7)}
& \makecell{4.364 (5.2) \\ 2.440 (24.2)}
& \makecell{3.766 (18.2) \\ 3.120 (3.1)}
& \makecell{4.668 (-1.4) \\ 2.636 (18.1)}
\\

CVRP & Serial + ME 
& \makecell{4.602 \\ 3.219}
& \makecell{4.221 (8.3) \\ 2.433 (24.4)}
& \makecell{\textbf{1.473 (68.0)} \\ \textbf{0.897 (72.1)}}
& \makecell{2.912 (36.7) \\ 1.829 (43.2)}
& \makecell{4.250 (7.7) \\ 2.501 (22.3)}
& \makecell{4.742 (-3.0) \\ 2.502 (22.3)}
& \makecell{3.128 (32.0) \\ 3.014 (6.4)}
& \makecell{4.602 (0.0) \\ 2.597 (19.3)}
\\

\bottomrule
\end{tabular}%
}
\vspace{0.25em}
\begin{minipage}{\textwidth}
\RaggedRight
\scriptsize
\textit{Notes.} Each component entry reports ``GAP (Gain\%)''. The first line corresponds to the 1,000-iteration budget, and the second line corresponds to the 60-second budget. Base GAP reports the average GAP of Baseline-ALNS under the two budgets. Gain is computed as $(\mathrm{GAP}_{base}-\mathrm{GAP}_{add})/\mathrm{GAP}_{base}\times100\%$. Bold values indicate the largest gain among the seven evolved components in the same row and budget. SE = single-expert; ME = multi-expert; D. Ctrl. = destruction-degree controller.
\end{minipage}
\end{threeparttable}
\end{table}

\begin{table}[htbp]
\centering
\caption{Subtractive ablation results across evolution paradigms}
\label{tab:subtractive_ablation_modes}
\tiny
\setlength{\tabcolsep}{2.2pt}
\renewcommand{\arraystretch}{1.15}
\begin{threeparttable}
\resizebox{\textwidth}{!}{%
\begin{tabular}{llllllllll}
\toprule
Problem & Mode & Full GAP & Destroy & Repair & Initial & Selector & Weight & Acceptance & D. Ctrl. \\
\midrule
TSP & Parallel + SE 
& \makecell{0.352 \\ 0.259}
& \makecell{0.384 (9.1) \\ 0.253 (-2.3)}
& \makecell{0.357 (1.4) \\ \textbf{0.761 (193.8)}}
& \makecell{0.361 (2.6) \\ 0.277 (7.0)}
& \makecell{0.388 (10.2) \\ 0.351 (35.5)}
& \makecell{0.357 (1.4) \\ 0.291 (12.4)}
& \makecell{0.366 (4.0) \\ 0.265 (2.3)}
& \makecell{\textbf{0.913 (159.4)} \\ 0.448 (73.0)}
\\

TSP & Parallel + ME 
& \makecell{0.428 \\ 0.395}
& \makecell{0.407 (-4.9) \\ 0.391 (-1.0)}
& \makecell{0.456 (6.5) \\ \textbf{0.682 (72.7)}}
& \makecell{0.480 (12.2) \\ 0.431 (9.1)}
& \makecell{0.530 (23.8) \\ 0.421 (6.6)}
& \makecell{0.463 (8.2) \\ 0.400 (1.3)}
& \makecell{0.426 (-0.5) \\ 0.399 (1.0)}
& \makecell{\textbf{0.625 (46.0)} \\ 0.194 (-50.9)}
\\

TSP & Serial + SE 
& \makecell{0.197 \\ 0.466}
& \makecell{0.353 (79.7) \\ 0.638 (36.9)}
& \makecell{0.306 (55.3) \\ 0.569 (22.1)}
& \makecell{0.235 (19.7) \\ 0.486 (4.3)}
& \makecell{0.214 (8.9) \\ 0.441 (-5.4)}
& \makecell{0.206 (4.8) \\ 0.483 (3.7)}
& \makecell{0.335 (70.4) \\ 0.525 (12.7)}
& \makecell{\textbf{0.881 (347.8)} \\ \textbf{0.752 (61.4)}}
\\

TSP & Serial + ME 
& \makecell{0.320 \\ 0.455}
& \makecell{1.014 (216.4) \\ \textbf{1.005 (120.9)}}
& \makecell{0.423 (32.2) \\ 0.591 (29.9)}
& \makecell{0.385 (20.2) \\ 0.568 (24.8)}
& \makecell{0.320 (0.0) \\ 0.469 (3.1)}
& \makecell{0.320 (0.0) \\ 0.456 (0.2)}
& \makecell{0.320 (0.0) \\ 0.463 (1.8)}
& \makecell{\textbf{1.080 (237.1)} \\ 0.711 (56.3)}
\\

\midrule

CVRP & Parallel + SE 
& \makecell{0.712 \\ 0.541}
& \makecell{0.815 (14.5) \\ 0.605 (11.8)}
& \makecell{\textbf{1.551 (118.0)} \\ \textbf{1.433 (164.9)}}
& \makecell{0.899 (26.4) \\ 0.484 (-10.5)}
& \makecell{0.727 (2.2) \\ 0.515 (-4.8)}
& \makecell{0.715 (0.5) \\ 0.478 (-11.7)}
& \makecell{1.060 (49.0) \\ 0.576 (6.5)}
& \makecell{0.673 (-5.4) \\ 0.468 (-13.5)}
\\

CVRP & Parallel + ME 
& \makecell{0.834 \\ 0.391}
& \makecell{0.864 (3.6) \\ 0.366 (-6.4)}
& \makecell{\textbf{1.475 (77.0)} \\ \textbf{0.797 (103.8)}}
& \makecell{1.019 (22.2) \\ 0.388 (-0.8)}
& \makecell{0.868 (4.1) \\ 0.352 (-10.0)}
& \makecell{0.892 (7.0) \\ 0.341 (-12.8)}
& \makecell{0.970 (16.3) \\ 0.486 (24.3)}
& \makecell{0.785 (-5.9) \\ 0.342 (-12.5)}
\\

CVRP & Serial + SE 
& \makecell{0.835 \\ 0.607}
& \makecell{1.108 (32.7) \\ 0.687 (13.2)}
& \makecell{\textbf{2.489 (198.1)} \\ \textbf{2.010 (231.1)}}
& \makecell{0.919 (10.1) \\ 0.588 (-3.1)}
& \makecell{0.951 (13.9) \\ 0.686 (13.0)}
& \makecell{1.025 (22.8) \\ 0.725 (19.4)}
& \makecell{0.907 (8.7) \\ 0.548 (-9.7)}
& \makecell{0.930 (11.4) \\ 0.621 (2.3)}
\\

CVRP & Serial + ME 
& \makecell{0.852 \\ 0.594}
& \makecell{1.252 (47.0) \\ 0.868 (46.1)}
& \makecell{\textbf{2.141 (151.4)} \\ \textbf{1.729 (191.1)}}
& \makecell{0.926 (8.7) \\ 0.636 (7.1)}
& \makecell{1.032 (21.1) \\ 0.818 (37.7)}
& \makecell{0.882 (3.6) \\ 0.643 (8.2)}
& \makecell{0.911 (7.0) \\ 0.595 (0.2)}
& \makecell{0.852 (0.0) \\ 0.592 (-0.3)}
\\

\bottomrule
\end{tabular}%
}
\vspace{0.25em}
\begin{minipage}{\textwidth}
\RaggedRight
\scriptsize
\textit{Notes.} Each component entry reports ``GAP (Loss\%)''. The first line corresponds to the 1,000-iteration budget, and the second line corresponds to the 60-second budget. Full GAP reports the average GAP of the fully evolved algorithm under the two budgets. Loss is computed as $(\mathrm{GAP}_{remove}-\mathrm{GAP}_{full})/\mathrm{GAP}_{full}\times100\%$. Bold values indicate the largest performance loss among the seven evolved components in the same row and budget. Negative values indicate that replacing the evolved component with its baseline counterpart slightly improves average performance. SE = single-expert; ME = multi-expert; D. Ctrl. = destruction-degree controller.
\end{minipage}
\end{threeparttable}
\end{table}

By analyzing the results presented in Table~\ref{tab:additive_ablation_modes} and Table~\ref{tab:subtractive_ablation_modes}, we summarize the following key findings:

\begin{itemize}
    \item \textbf{Component contributions are highly problem-specific:} 
    \begin{itemize}
        \item \textit{For the TSP:} The additive ablation (Table~\ref{tab:additive_ablation_modes}) demonstrates that evolving the \textbf{Destroy operators} yields the highest standalone performance gain, improving the baseline by approximately 62\% to 76\%. Conversely, the subtractive ablation (Table~\ref{tab:subtractive_ablation_modes}) reveals that the Destruction-degree controller (D. Ctrl.) plays a crucial foundational role in the fully synergistic system; its removal leads to the most severe performance degradation, with losses reaching up to 347.8\% in certain modes. This suggests that the bottleneck in solving the TSP primarily lies in effectively identifying ``how to break the existing solution'' and ``how to control the magnitude of disruption.''
        
        \item \textit{For the CVRP:} The Repair operators exhibit absolute dominance. They not only provide the highest standalone gain in the additive ablation (approximately 67\% to 76\%) but also cause catastrophic performance drops (typically ranging from 77\% to 231.1\%) when removed in the subtractive ablation. This confirms that, given the complex capacity constraints of the CVRP, the ability to ``feasibly and efficiently reconstruct the solution'' is the core element for breaking through the algorithm's performance bottleneck.
    \end{itemize}
    
    \item \textbf{Robustness and consistency across paradigms:} 
    Although different evolution paradigms (e.g., SE vs.\ ME, Parallel vs.\ Serial) yield varying overall GAP performances, the core attribution patterns, TSP prioritizing destruction and its control, while CVRP prioritizes repair, remain remarkably consistent across all tested modes and computational budgets. This empirical evidence demonstrates that the LLM-driven evolution process does not merely perform blind optimizations; rather, it adaptively pinpoints and optimizes the distinct structural bottlenecks of different combinatorial optimization problems.
\end{itemize}

\subsection{Does the Choice of Underlying LLM Matter?}
\label{sec:llm}

This subsection examines whether the choice of the underlying LLM affects the quality of the evolved ALNS.
We use the TSP as the controlled testbed for this analysis, and for each LLM, we run a complete full-component evolution process under the same parallel single-expert setting.

\subsubsection{Cross-LLM Performance Profile}

The code generation and reasoning ability of the LLM used as the evolutionary engine significantly influence the quality of the resulting algorithm. To investigate model-specific performance, three LLMs are compared: \texttt{DeepSeek-V3.2}, \texttt{Grok-4.1-Fast}, and \texttt{Gemini-3-Flash}. Table~\ref{tab:multi-llm} reports the average GAP across different budget settings and instance groups.

\begin{table}[thbp]
\centering
\scriptsize
\caption{Performance comparison of evolved algorithms driven by different LLMs (average GAP).}
\label{tab:multi-llm}
\begin{tabular}{llcccc}
\hline
\textbf{Setting} & \textbf{Dataset split} & Baseline & \texttt{DeepSeek-V3.2} & \texttt{Grok-4.1-Fast} & \texttt{Gemini-3-Flash} \\
\hline
\multirow{4}{*}{1,000 iterations}
& Small-Evo      & 1.714\% & 0.365\% & 0.402\% & \textbf{0.239\%} \\
& Med-Evo        & 2.007\% & 0.688\% & 0.803\% & \textbf{0.485\%} \\
& MS-Test        & 2.749\% & \textbf{1.170\%} & 1.425\% & 1.230\% \\
& Lg-Test        & 4.177\% & 2.089\% & 2.255\% & \textbf{1.844\%} \\
\hline
\multirow{4}{*}{Runtime 60s}
& Small-Evo      & 1.662\% & 0.413\% & \textbf{0.058\%} & 0.098\% \\
& Med-Evo        & 3.723\% & 1.404\% & 0.474\% & \textbf{0.449\%} \\
& MS-Test        & 6.135\% & 2.429\% & \textbf{1.118\%} & 1.138\% \\
& Lg-Test        & 10.378\%& 4.708\% & 2.723\% & \textbf{2.680\%} \\
\hline
\multirow{4}{*}{10,000 iterations}
& Small-Evo      & 0.300\% & 0.050\% & 0.078\% & \textbf{0.047\%} \\
& Med-Evo        & 1.071\% & 0.202\% & 0.258\% & \textbf{0.161\%} \\
& MS-Test        & 1.326\% & \textbf{0.461\%} & 0.523\% & 0.493\% \\
& Lg-Test        & 3.251\% & 1.212\% & 1.293\% & \textbf{1.026\%} \\
\hline
\end{tabular}
\end{table}

From Table~\ref{tab:multi-llm}, the following observations can be drawn:

\begin{enumerate}
  \item \textbf{Fixed 1,000 iterations: Gemini-3-Flash yields the highest search depth.} 
  Under the same iteration budget, the algorithm evolved by \texttt{Gemini-3-Flash} achieves the lowest average GAP in three out of four instance groups. This suggests that its operator logic and control strategies are highly effective per search step, although \texttt{DeepSeek-V3.2} shows a slight advantage on the medium-small test instances.

  \item \textbf{Fixed 60 seconds: Grok-4.1-Fast and Gemini-3-Flash demonstrate superior efficiency.} 
  The performance gap between models becomes more pronounced under strict time constraints. \texttt{Grok-4.1-Fast} and \texttt{Gemini-3-Flash} achieve significantly lower GAPs compared to \texttt{DeepSeek-V3.2}. Notably, \texttt{Grok-4.1-Fast} excels on small and medium-small instances, while \texttt{Gemini-3-Flash} maintains the lead on medium and large instances, indicating that both models are capable of generating low-overhead, high-throughput heuristics.

  \item \textbf{Fixed 10,000 iterations: Gemini-3-Flash and DeepSeek-V3.2 show robust long-run convergence.} 
  Under an extended iteration budget, \texttt{Gemini-3-Flash} and \texttt{DeepSeek-V3.2} emerge as the strongest competitors. While \texttt{DeepSeek-V3.2} performs exceptionally well on the medium-small categories, \texttt{Gemini-3-Flash} provides the best global results for other instances. This suggests that the algorithms evolved by Gemini-3-Flash and DeepSeek-V3.2 remain more competitive under extended iteration budgets.
\end{enumerate}

Overall, Table~\ref{tab:multi-llm} shows that, regardless of which LLM is used as the evolutionary engine, the resulting Evolved-ALNS consistently outperforms \textit{Baseline-ALNS} across all evaluation settings, confirming the effectiveness of LLM-guided evolution for improving ALNS design. At the same time, the three LLMs exhibit clearly different performance profiles rather than behaving as interchangeable optimizers: \texttt{Gemini-3-Flash} is generally strongest under fixed iteration budgets, \texttt{Grok-4.1-Fast} is especially competitive under strict runtime limits, and \texttt{DeepSeek-V3.2} remains highly competitive on some held-out test settings, especially the medium-small test subset. This suggests that the choice of LLM should be aligned with the primary objective of the evolutionary process and budget.

\subsection{Does Cross-Problem Transfer Preserve the Effectiveness of Evolved ALNS?}
\label{sec:transfer}

To evaluate the generalization capability of the evolved algorithmic logic, a cross-problem transferability experiment was conducted. Specifically, the high-level decision and control components (operator selector, weight updater, destruction degree controller, and acceptance criterion) were swapped between TSP and CVRP. The objective is to determine if the LLM-driven evolution discovers "problem-agnostic" meta-heuristic principles that remain effective beyond their original training domain.

Table~\ref{tab:transfer_vs_baseline_native} shows a clear but asymmetric transfer pattern. When high‑level control components are moved from CVRP to TSP, the transferred policy remains competitive with the native TSP solver while still improving substantially over the baseline. Under 1,000 iterations, the transfer is slightly worse than the native policy in three of the four evolution modes ($\Delta$GAP ranging from +16.7\% to +67.6\%), but outperforms it in Parallel + ME (-12.6\%); under 10,000 iterations, the gap is mixed, with one mode improves, two show moderate degradation, and Serial + ME degrades substantially. In all cases, runtime is significantly reduced, often by more than 40\%. Under the 60‑second budget, three of the four modes show improved or comparable GAP while completing many more iterations, confirming the practical efficiency of the CVRP‑evolved control logic on TSP.

\begin{table}[!htbp]
\centering
\caption{Cross-problem transfer performance under four evolution modes, averaged over all benchmark instances.}
\label{tab:transfer_vs_baseline_native}
\scriptsize
\begin{threeparttable}
\resizebox{\linewidth}{!}{%
\begin{tabular}{lllllll}
\toprule
\textbf{Direction} 
& \textbf{Evolution mode}
& \textbf{Baseline}
& \textbf{Native}
& \textbf{Transfer}
& \makecell{\textbf{Transfer vs. Native}\\ \textbf{$\Delta$GAP / $\Delta$Effort}}
& \makecell{\textbf{Transfer vs. Baseline}\\ \textbf{GAP improvement / $\Delta$Effort}} \\
\midrule

\multicolumn{7}{l}{\textit{Panel A: Fixed iterations = 1,000; entries are mean GAP / Time(s)}} \\
CVRP $\rightarrow$ TSP & Parallel + SE & 2.438 / 962.78 & 0.816 / 51.76 & 0.952 / 29.22 & +16.7 / -43.6 & +61.0 / -97.0 \\
CVRP $\rightarrow$ TSP & Parallel + ME & 2.438 / 962.78 & 0.908 / 154.98 & 0.794 / 36.29 & -12.6 / -76.6 & +67.4 / -96.2 \\
CVRP $\rightarrow$ TSP & Serial + SE & 2.438 / 962.78 & 0.561 / 944.62 & 0.861 / 590.81 & +53.6 / -37.5 & +64.7 / -38.6 \\
CVRP $\rightarrow$ TSP & Serial + ME & 2.438 / 962.78 & 0.815 / 952.17 & 1.366 / 539.00 & +67.6 / -43.4 & +44.0 / -44.0 \\
TSP $\rightarrow$ CVRP & Parallel + SE & 6.328 / 59.69 & 1.046 / 55.96 & 2.315 / 109.47 & +121.2 / +95.6 & +63.4 / +83.4 \\
TSP $\rightarrow$ CVRP & Parallel + ME & 6.328 / 59.69 & 1.161 / 29.53 & 3.014 / 130.87 & +159.7 / +343.2 & +52.4 / +119.3 \\
TSP $\rightarrow$ CVRP & Serial + SE & 6.328 / 59.69 & 1.344 / 38.49 & 2.024 / 113.25 & +50.6 / +194.2 & +68.0 / +89.7 \\
TSP $\rightarrow$ CVRP & Serial + ME & 6.328 / 59.69 & 1.229 / 50.08 & 1.993 / 129.62 & +62.1 / +158.9 & +68.5 / +117.2 \\
\addlinespace[2pt]

\multicolumn{7}{l}{\textit{Panel B: Fixed iterations = 10,000; entries are mean GAP / Time(s)}} \\
CVRP $\rightarrow$ TSP & Parallel + SE & 1.217 / 9354.90 & 0.344 / 794.08 & 0.408 / 273.10 & +18.5 / -65.6 & +66.5 / -97.1 \\
CVRP $\rightarrow$ TSP & Parallel + ME & 1.217 / 9354.90 & 0.413 / 1343.72 & 0.408 / 245.32 & -1.2 / -81.7 & +66.5 / -97.4 \\
CVRP $\rightarrow$ TSP & Serial + SE & 1.217 / 9354.90 & 0.349 / 12222.03 & 0.432 / 5952.81 & +23.9 / -51.3 & +64.5 / -36.4 \\
CVRP $\rightarrow$ TSP & Serial + ME & 1.217 / 9354.90 & 0.268 / 9147.81 & 0.640 / 5293.45 & +138.9 / -42.1 & +47.4 / -43.4 \\
TSP $\rightarrow$ CVRP & Parallel + SE & 2.503 / 690.93 & 0.576 / 716.62 & 1.818 / 924.83 & +215.7 / +29.1 & +27.4 / +33.9 \\
TSP $\rightarrow$ CVRP & Parallel + ME & 2.503 / 690.93 & 0.479 / 516.32 & 1.498 / 898.42 & +212.6 / +74.0 & +40.1 / +30.0 \\
TSP $\rightarrow$ CVRP & Serial + SE & 2.503 / 690.93 & 0.662 / 436.10 & 1.843 / 1396.03 & +178.3 / +220.1 & +26.4 / +102.1 \\
TSP $\rightarrow$ CVRP & Serial + ME & 2.503 / 690.93 & 0.601 / 523.38 & 1.010 / 1406.44 & +68.0 / +168.7 & +59.6 / +103.6 \\
\addlinespace[2pt]

\multicolumn{7}{l}{\textit{Panel C: Fixed time = 60 seconds; entries are mean GAP / Iterations}} \\
CVRP $\rightarrow$ TSP & Parallel + SE & 4.725 / 462.10 & 0.856 / 1849.76 & 0.760 / 2867.57 & -11.2 / +55.0 & +83.9 / +520.6 \\
CVRP $\rightarrow$ TSP & Parallel + ME & 4.725 / 462.10 & 1.210 / 1588.45 & 0.942 / 3726.38 & -22.2 / +134.6 & +80.1 / +706.4 \\
CVRP $\rightarrow$ TSP & Serial + SE & 4.725 / 462.10 & 1.296 / 244.59 & 1.375 / 403.35 & +6.1 / +64.9 & +70.9 / -12.7 \\
CVRP $\rightarrow$ TSP & Serial + ME & 4.725 / 462.10 & 1.602 / 435.17 & 1.999 / 1267.51 & +24.8 / +191.3 & +57.7 / +174.3 \\
TSP $\rightarrow$ CVRP & Parallel + SE & 5.304 / 4810.61 & 0.985 / 2362.62 & 2.348 / 2237.15 & +138.5 / -5.3 & +55.7 / -53.5 \\
TSP $\rightarrow$ CVRP & Parallel + ME & 5.304 / 4810.61 & 0.900 / 2796.45 & 2.843 / 2110.99 & +215.7 / -24.5 & +46.4 / -56.1 \\
TSP $\rightarrow$ CVRP & Serial + SE & 5.304 / 4810.61 & 1.259 / 3741.95 & 2.492 / 1446.28 & +97.9 / -61.3 & +53.0 / -69.9 \\
TSP $\rightarrow$ CVRP & Serial + ME & 5.304 / 4810.61 & 1.124 / 2758.84 & 2.157 / 1039.26 & +91.9 / -62.3 & +59.3 / -78.4 \\
\bottomrule
\end{tabular}%
}

\begin{tablenotes}[flushleft]
\tiny
\item Note. All entries are averaged over all benchmark instances for the corresponding evolution mode. GAP values are reported in percentage points. ``Native'' denotes the Evolved-Gemini algorithm directly evolved on the target problem, whereas ``Transfer'' denotes the configuration using high-level control components transferred from the source problem. In Panels A and B, effort is measured by runtime in seconds; in Panel C, effort is measured by the number of iterations completed within 60 seconds. For ``Transfer vs. Native,'' $\Delta$GAP is computed as $(\mathrm{GAP}_{\mathrm{Transfer}}-\mathrm{GAP}_{\mathrm{Native}})/\mathrm{GAP}_{\mathrm{Native}}\times100\%$. For ``Transfer vs. Baseline,'' GAP improvement is computed as $(\mathrm{GAP}_{\mathrm{Baseline}}-\mathrm{GAP}_{\mathrm{Transfer}})/\mathrm{GAP}_{\mathrm{Baseline}}\times100\%$. For effort changes, negative values in Panels A and B indicate shorter runtime, whereas positive values in Panel C indicate more iterations completed within the fixed time budget.
\end{tablenotes}
\end{threeparttable}
\end{table}

The reverse transfer, from TSP to CVRP, is less effective and less robust. In all evolution modes and budgets, the transferred policy substantially degrades relative to the native CVRP solver, with $\Delta$GAP increases ranging from +50.6\% to +159.7\% under 1,000 iterations. Although the transferred configuration still outperforms the baseline ALNS in most settings, the margin is considerably smaller, and computational effort can be higher. These results reinforce that the evolved high‑level ALNS logic is partially reusable across problems, but the transferability is asymmetric: control policies evolved on the more constrained CVRP generalize reasonably well to TSP, whereas those evolved on TSP do not transfer effectively to the richer constraint structure of CVRP.



\subsection{\textcolor{black}{How Does Evolved-ALNS Compare with Other LLM-Driven and Learning-Based Solvers?}}
\label{sec:vs-evolved}

To place the evolved ALNS in a broader methodological context, we benchmark it against
two external solvers: Generative Large Neighborhood Search (G-LNS) \citep{zhao2026g}, a learning-augmented LNS, and Evolution of Heuristic Sets (EoH-S) \citep{liu2026eoh}, a
recent LLM-driven heuristic-evolution approach. All algorithms are evaluated on the overlapping instance pool shared by these two studies and this paper (40 TSP and 47 CVRP instances).
For each problem, we report Parallel~+~SE together with a serial variant (Serial~+~ME for TSP, Serial~+~SE for CVRP).
Table~\ref{tab:cross_solver} summarizes the results.

\begin{table}[htbp]
\centering
\caption{Comparison of Evolved-ALNS with G-LNS and EoH-S.
Values are optimality gaps (\%); lower is better. ``Best~\#'' counts instances where
the method attains the strictly lowest gap among the four solvers.}
\label{tab:cross_solver}
\small
\setlength{\tabcolsep}{6pt}
\begin{tabular}{llcccccc}
\toprule
Problem & Method & Mean & Standard deviation & Minimum & Median & Maximum & Best~\# \\
\midrule
\multirow{4}{*}{\shortstack[l]{TSP\\($n=40$)}}
 & Parallel + SE & 0.95 & 0.88 & 0.00 & 0.73 & 2.80 & 25 \\
 & Serial + ME   & 1.00 & 0.95 & 0.00 & 0.74 & 3.24 & 13 \\
 & G-LNS         & 2.37 & 1.46 & 0.00 & 2.50 & 6.80 & 2  \\
 & EoH-S         & 8.46 & 3.39 & 3.40 & 8.50 & 16.90 & 0 \\
\midrule
\multirow{4}{*}{\shortstack[l]{CVRP\\($n=47$)}}
 & Parallel + SE & 1.34 & 0.95 & 0.00 & 1.31 & 3.54 & 36 \\
 & Serial + SE   & 1.77 & 1.29 & 0.00 & 1.61 & 5.08 & 11 \\
 & G-LNS         & 9.31 & 4.07 & 2.30 & 8.20 & 18.90 & 0  \\
 & EoH-S         & 21.60 & 7.65 & 6.50 & 20.10 & 39.40 & 0 \\
\bottomrule
\end{tabular}

\vspace{2pt}
\begin{minipage}{\linewidth}
\footnotesize
\textit{Notes.} Parallel~+~SE, Serial~+~ME / Serial~+~SE, and G-LNS are run under a fixed
budget of 500 iterations. EoH-S is a constructive heuristic set (no iterative search);
its solutions are re-evaluated by G-LNS under the original coordinates as a single
evaluation, whereas the original EoH-S operates on normalized coordinates.
\end{minipage}
\end{table}

Under our unified evaluation protocol, the proposed method achieves lower gaps than both external solvers across both problems. Parallel~+~SE reduces the mean gap from 2.37\% (G-LNS) and 8.46\% (EoH-S) to 0.95\% on TSP, and from
9.31\% and 21.60\% to 1.34\% on CVRP. The advantage is uniform: the maximum gap of
Parallel~+~SE is below the \emph{minimum} gap of EoH-S on both problems, and the evolved
variants attain the strictly best gap on 85 of the 87 instances. Notably, EoH-S also
relies on LLM-driven evolution but only at the operator level; the resulting gap relative
to our framework supports the central claim that full-component evolution is needed to
fully exploit the LLM as a code-generation engine.

\subsection{Discussion: Scientific Insights}
\label{sec:discussion}

\subsubsection{A note on multi-expert evolution: when component-level gains fail to compose}

A particularly informative pattern emerges when the parallel rows in Table~\ref{tab:additive_ablation_modes} are compared with the integrated results in Table~\ref{tab:tsp_cvrp_evolution_modes}. At the component level, the multi-expert prompt is usually stronger. In the additive ablation under parallel evolution, \textit{Parallel + ME} achieves a higher Gain\% than \textit{Parallel + SE} in 23 of the 28 component--budget comparisons across TSP and CVRP. Broken down by problem, the multi-expert setting is better in 13 of the 14 TSP comparisons and 10 of the 14 CVRP comparisons. If one looked only at the ablation table, the natural expectation would therefore be that multi-expert evolution should also produce the stronger full ALNS system.

However, Table~\ref{tab:tsp_cvrp_evolution_modes} shows that this inference does not reliably hold once the evolved modules are assembled. The contrast is sharpest on TSP. Although the parallel multi-expert setting produces better standalone modules in the ablation study, the final \textit{Parallel + SE} solver outperforms \textit{Parallel + ME} on all four dataset splits under all three evaluation budgets. In other words, the multi-expert advantage is local, but not compositional. On CVRP, the same phenomenon is weaker rather than absent: \textit{Parallel + ME} remains competitive, and even overtakes \textit{Parallel + SE} under 10{,}000 iterations and on three of the four splits under the 60-second budget, yet its broader component-level advantage still does not translate into a uniformly dominant integrated system. Across both problems, the more general lesson is that better modules do not necessarily sum to a better algorithm.

This mismatch suggests that parallel multi-expert evolution may improve modules along partially different inductive biases, yielding components that are individually strong but less mutually compatible when combined. By contrast, single-expert evolution may produce modules that are slightly less impressive in isolation yet more stylistically and strategically coherent as a set. Because the parallel pipeline evolves each component largely against fixed sparring partners, it is especially vulnerable to this composition gap: a module can appear strong in isolation, yet still fail to match the style, search logic, or implicit assumptions of other evolved modules, especially when those modules are themselves produced under substantially different prompt personas. It implies that the main difficulty in automated metaheuristic design is not only discovering stronger local components, but also preserving cross-component compatibility so that local gains survive system-level integration.

\subsubsection{Insights from the Evolved Code}

Analysis of the evolved ALNS code for TSP and CVRP shows that its performance improvement does not come from isolated modifications to a single component. Instead, it arises from coordinated changes across search control, perturbation design, and adaptive decision-making. Overall, the evolution process tends to transform the relatively static and experience-driven rules of traditional ALNS into mechanisms that are more scale-adaptive, state-aware, and diversity-preserving.

First, the evolved code strengthens the dynamic control of the search process. Acceptance criteria often replace absolute objective differences with relative deterioration measures, improving stability across instances of different scales. Similarly, destruction-degree controllers no longer follow fixed ratios or monotone decay schedules. Instead, they combine search progress, stagnation status, oscillatory perturbations, and occasional large-scale destructions to switch more flexibly between intensification and escape. This suggests that effective ALNS control does not necessarily require smooth convergence; it can also be non-monotone, stage-dependent, and restart-like.

Second, the evolved code introduces stronger exploration-preserving mechanisms in operator selection and weight updating. Compared with classical roulette-wheel selection and fixed exponential smoothing, many evolved variants use softmax sampling, upper-confidence-bound (UCB)-like bonuses, success-rate priors, explicit exploration floors, and weight bounds. These mechanisms prevent a few high-scoring operators from monopolizing the search too early. As a result, ALNS adaptation is better interpreted as an online exploration–exploitation problem rather than a simple accumulation of historical scores.

Third, the evolved destroy and repair operators show stronger awareness of problem structure. For TSP, destroy operators tend to remove spatially related or structurally inefficient tour segments. For CVRP, destroy and repair logic further incorporates demand, capacity, and depot-distance information. In particular, CVRP repair operators often prioritize high-demand or depot-distant customers, reducing the feasibility risk of later insertions. This indicates that effective ALNS operators should not only respond to local cost changes, but also explicitly exploit structural features of the underlying problem.

Finally, a recurring counterintuitive observation is that the evolved code does not pursue fully deterministic greedy search. Instead, controlled noise, randomized insertion, Greedy Randomized Adaptive Search Procedure (GRASP)-like construction, and multi-start screening are widely used in initial solution generation and repair. These stochastic mechanisms do not simply weaken solution quality; rather, they reduce the risk of repeatedly reconstructing similar solution structures and improve robustness through controlled diversification.

In summary, the key outcome of LLM-driven evolution is not a single optimal formula, but a set of coordinated design principles: scale-normalized acceptance, non-monotone perturbation control, diversity-preserving adaptive learning, structure-aware destroy and repair, and moderately randomized reconstruction. These mechanisms suggest that LLM-driven program evolution can improve ALNS not only computationally, but also scientifically, by revealing algorithmic design patterns that are difficult to identify through conventional manual design.

All code generated through the four evolutionary modes can be downloaded at the following URL: https://github.com/Shaohua-Yu/LLM-ALNS.

\section{Conclusion}

This paper studied whether large language models can move beyond operator-level assistance and support the full-component design of Adaptive Large Neighborhood Search. To that end, we proposed a closed-loop evolutionary framework that decomposes ALNS into seven modules and evolves them under controlled evaluation, while MAP-Elites preserves strategically diverse high-performing candidates throughout the search. The resulting framework shifts ALNS development from manual component crafting toward a more systematic search over complete algorithmic designs.

The computational results show that this full-component perspective matters. On both TSP and CVRP, the evolved algorithms consistently outperform the optimized classical ALNS baselines under fixed-iteration and fixed-time budgets, and the advantages remain visible as instance size increases. The ablation experiments further show that performance gains do not arise from a single dominant innovation. Instead, the strongest improvements come from coordinated changes across multiple layers of the method, especially the destroy--repair machinery and its associated control logic, which confirms that ALNS performance is fundamentally an emergent property of interdependent components rather than a simple sum of isolated module improvements.

Beyond aggregate performance, the study also provides several substantive insights into LLM-driven algorithm design. First, the evolved code reveals non-intuitive but effective mechanisms that would be difficult to obtain through conventional manual tuning alone, suggesting that LLM-guided evolution can serve not only as an optimization tool but also as a discovery tool. Second, the comparison across model families shows that the choice of LLM materially affects the quality and style of the evolved algorithms, implying that model selection is itself a design decision in automated heuristic generation. Third, the cross-problem transfer experiments indicate that part of the evolved high-level logic is reusable across domains, but transferability is asymmetric: control policies evolved on CVRP transfer to TSP with competitive performance and substantial runtime savings, whereas those evolved on TSP do not generalize well to CVRP. This result suggests that LLM evolution can uncover reusable meta-heuristic principles, while also highlighting the limits of their portability.

Taken together, these findings suggest that the main value of LLM-driven ALNS evolution lies not merely in automating coding effort, but in enlarging the space of algorithmic ideas that can be searched, tested, and recombined. At the same time, the results also expose important limitations. The current framework still relies on expensive empirical evaluation. In practice, algorithm evolution itself becomes a form of expensive optimization: evolving even a single component over 300 generations already requires hundreds of repeated executions of the full heuristic pipeline, together with substantial computational time and LLM token budget. Although the present framework already borrows several practical ideas from expensive optimization to improve throughput, the overall search remains costly. In addition, the module-wise decomposition, while computationally necessary, cannot fully capture compatibility effects among strongly coupled components. These observations point to several promising directions for future research, including more theory-guided and sample-efficient evolution schemes, co-evolution of interacting modules, lower-cost and more reliable surrogate evaluation, and extension to richer routing and scheduling problems with more complex feasibility structures.

Overall, the evidence in this paper supports a simple conclusion: LLMs can contribute to heuristic design not only by generating local operators, but by helping evolve coherent, high-performing metaheuristic systems. We view this as a step toward a more general paradigm in which large models act as partners in scientific algorithm design, combining automated exploration with interpretable algorithmic insight.

\bigskip

\section*{Data availability statement}

The ALNS code generated by different LLMs under different evolutionary modes in the paper can be downloaded from the following URL: https://github.com/Shaohua-Yu/LLM-ALNS.

\section*{CRediT authorship contribution statement}

Shaohua Yu: Conceptualization, Methodology, Software, Formal analysis, Resources, Writing – original draft, Supervision, Funding acquisition.
Tianyu Chen: Methodology, Software, Investigation, Data curation, Writing – original draft.
Linyan Liu: Validation, Investigation, Supervision.
Jakob Puchinger: Methodology, Validation, Writing – original draft.

\section*{Acknowledgements}

The research is supported by National Natural Science Foundation of China [grant no. 72301137].


\bibliographystyle{apalike}
\bibliography{sample}

\appendix
\section{Benchmark instance lists}
\label{A:1}
This appendix provides the complete list of TSPLIB and CVRPLIB instances used in the computational experiments reported in Section 4. The instances are organized according to the evolution set / test set partition and the size-based grouping.

\subsection{TSP instances (TSPLIB)}

For TSP, the size categories are defined continuously as follows: small-scale instances have $n \le 105$, medium-scale instances have $106 \le n < 300$, and large-scale instances have $n \ge 300$.

Evolution set — small-scale subset (13 instances, $n \le 105$). \texttt{oliver30}, \texttt{berlin52}, \texttt{eil51}, \texttt{eil76}, \texttt{eil101}, \texttt{kroA100}, \texttt{kroC100}, \texttt{kroD100}, \texttt{kroE100}, \texttt{pr76}, \texttt{rat99}, \texttt{st70}, \texttt{lin105}.

Evolution set — medium-scale subset (11 instances, $106 \le n < 300$). \texttt{bier127}, \texttt{ch130}, \texttt{ch150}, \texttt{kroA150}, \texttt{kroA200}, \texttt{kroB150}, \texttt{kroB200}, \texttt{pr124}, \texttt{pr136}, \texttt{pr152}, \texttt{pr226}.

Test set — small- and medium-scale subset (12 instances, $n < 300$).  \texttt{a280}, \texttt{d198}, \texttt{gil262},  \texttt{pr107}, \texttt{pr144}, \texttt{pr264}, \texttt{pr299}, \texttt{rat195}, \texttt{rd100}, \texttt{ts225}, \texttt{tsp225}, \texttt{u159}.

Test set — large-scale subset (6 instances, $n \ge 300$).  \texttt{d493}, \texttt{fl417}, \texttt{lin318}, \texttt{linhp318}, \texttt{pr439}, \texttt{rd400}.

\subsection{CVRP instances (CVRPLIB)}

For CVRP, the size categories are defined continuously as follows: small-scale instances have $n < 60$, medium-scale instances have $60 \le n < 200$, and large-scale instances have $n \ge 200$.

Evolution set — small-scale subset (12 instances, $n < 60$). \texttt{E-n22-k4}, \texttt{P-n23-k8}, \texttt{B-n31-k5}, \texttt{A-n32-k5}, \texttt{E-n33-k4}, \texttt{A-n37-k5}, \texttt{B-n38-k6}, \texttt{A-n45-k6}, \texttt{B-n45-k5}, \texttt{E-n51-k5}, \texttt{A-n53-k7}, \texttt{P-n50-k10}.

Evolution set — medium-scale subset (18 instances, $60 \le n < 200$). \texttt{A-n65-k9}, \texttt{A-n69-k9}, \texttt{A-n80-k10}, \texttt{B-n66-k9}, \texttt{B-n67-k10}, \texttt{B-n68-k9}, \texttt{B-n78-k10}, \texttt{E-n76-k7}, \texttt{E-n76-k8}, \texttt{E-n76-k10}, \texttt{E-n76-k14}, \texttt{E-n101-k8}, \texttt{E-n101-k14}, \texttt{P-n60-k10}, \texttt{P-n70-k10}, \texttt{P-n76-k4}, \texttt{P-n76-k5}, \texttt{P-n101-k4}.

Test set — small- and medium-scale subset (15 instances, $n < 200$). \texttt{A-n54-k7}, \texttt{A-n60-k9}, \texttt{A-n62-k8}, \texttt{A-n63-k9}, \texttt{A-n64-k9}, \texttt{P-n55-k10}, \texttt{P-n60-k15}, \texttt{P-n65-k10}, \texttt{P-n51-k10}, \texttt{X-n101-k25}, \texttt{X-n106-k14}, \texttt{X-n110-k13}, \texttt{X-n115-k10}, \texttt{X-n120-k6}, \texttt{X-n129-k18}.

Test set — large-scale subset (5 instances, $n \ge 200$). \texttt{X-n200-k36}, \texttt{X-n204-k19}, \texttt{X-n214-k11}, \texttt{X-n228-k23}, \texttt{X-n233-k16}.

\end{document}